\documentclass[11pt]{article}

\usepackage[utf8]{inputenc}
\usepackage[T1]{fontenc}

\usepackage{arxiv}

\usepackage{natbib}
\usepackage{url}
\usepackage{booktabs}
\usepackage{amsmath,amssymb}
\usepackage{graphicx}
\usepackage{float}
\usepackage{xcolor}
\usepackage{multirow}
\usepackage{array}
\usepackage{enumitem}
\usepackage{algorithm}
\usepackage{algorithmic}
\usepackage{subcaption}
\usepackage{tablefootnote}
\usepackage{colortbl}
\usepackage{pifont}  
\usepackage{tikz}
\usetikzlibrary{arrows.meta,positioning,fit,backgrounds,calc,decorations.pathreplacing,patterns,shapes.geometric}
\usepackage{pgfplots}
\pgfplotsset{compat=1.18}
\usepgfplotslibrary{groupplots,fillbetween}
\usepackage{tcolorbox}
\tcbuselibrary{skins,breakable}

\usepackage{hyperref}
\hypersetup{
  colorlinks=true,
  linkcolor=arxivaccent,
  citecolor=arxivaccent,
  urlcolor=arxivaccent,
  filecolor=arxivaccent,
  breaklinks=true,
}

\setlist{nosep,leftmargin=*}

\definecolor{bestcolor}{HTML}{E8F5E9}
\definecolor{gaincolor}{HTML}{1B5E20}
\definecolor{cStandard}{HTML}{BF616A}   
\definecolor{cAOI}{HTML}{5E81AC}        
\definecolor{cDelta}{HTML}{A3BE8C}      
\definecolor{cVisual}{HTML}{D08770}     
\definecolor{cASR}{HTML}{B48EAD}        
\definecolor{cNarr}{HTML}{88C0D0}       
\definecolor{cAccent}{HTML}{EBCB8B}     
\definecolor{cGray}{HTML}{4C566A}       
\definecolor{cLightGray}{HTML}{D8DEE9}  

\title{Agent-Computer Observation Interfaces\\Enable Dynamic Computer Use}

\author{%
  \begin{tabular}{@{}c@{\hspace{4em}}c@{}}
    Bojie Li & Noah Shi \\
    Pine AI & University of Washington
  \end{tabular}%
}
\date{}
\runningtitle{Agent-Computer Observation Interfaces Enable Dynamic Computer Use}

\begin{document}

\maketitle

\begin{abstract}
SWE-agent established the \emph{action} interface as an underexplored design axis for software-engineering agents; we make the analogous case for the \emph{observation} interface in computer-use (CU) agents. Current CU agents, closed and open-source alike, tie observation to action---one screenshot every 3--5\,s, no audio---leaving them blind and deaf between screenshots to video, animations, transient UI events, meetings, and spoken instructions. We introduce the \textbf{Agent-Computer Observation Interface (AOI)}, a model-agnostic perception layer that decouples continuous, adaptive observation from discrete actions through three gated components: inter-step keyframe capture, volume-gated audio transcription, and CU-model-generated visual narration that persists as text. Each produces almost nothing on static, silent content, reducing to the standard loop without degrading it.

On \textbf{DynaCU-Bench} (100 dynamic browser tasks plus a 50-task static control), CU models from 7B to frontier scale gain $+17$ to $+48$\,pp over their screenshot baselines with zero retraining, turning tasks that are near-impossible from periodic screenshots into largely solved ones. The gap is starkest on audio: on a spoken-content subset AOI agents solve every task, whereas streaming voice models hear accurately but cannot act on what they hear without the scaffold. The decomposition is as informative as the headline gain: keyframe \emph{selection} turns out not to matter---the value comes from narrating captured frames into persistent text---and the interface is not a fixed bundle, since on a newer model (Gemini~3~Flash) the keyframe stream actively regresses through image-token dilution, so its components must be selected per model rather than shipped as one configuration.
\end{abstract}

\begin{center}
\small
Code: \url{https://github.com/19PINE-AI/aoi} \quad\textbar\quad Website: \url{https://01.me/research/aoi}
\end{center}
\vspace{-0.6em}

\begin{figure}[H]
\centering
\begin{tikzpicture}
\begin{axis}[
    width=\columnwidth,
    height=4.7cm,
    ybar,
    bar width=6.5pt,
    ymin=0, ymax=112,
    ylabel={Task Success (\%)},
    ylabel style={font=\small},
    symbolic x coords={Claude 4.6, GPT-5.4, Gemini 2.5, Gemini 3, Grok-4, Grok-4.3, Grok-4-fast, EvoCUA-32B, Fara-7B},
    xtick=data,
    xticklabel style={font=\tiny, rotate=30, anchor=east},
    ytick={0,20,40,60,80,100},
    yticklabel style={font=\small},
    legend style={at={(0.98,1.0)}, anchor=north east, font=\scriptsize,
                  legend columns=2, /tikz/every even column/.append style={column sep=4pt},
                  draw=cGray!30, fill=white, fill opacity=0.9},
    grid=major,
    major grid style={cLightGray!60},
    axis line style={cGray},
    tick style={cGray},
    enlarge x limits=0.06,
    nodes near coords,
    nodes near coords style={font=\tiny\bfseries, /pgf/number format/fixed},
    every node near coord/.append style={above=1pt},
    clip=false,
]
\addplot[fill=cStandard!70, draw=cStandard!90]
    coordinates {
      (Claude 4.6,    38)
      (GPT-5.4,       37)
      (Gemini 2.5,    21)
      (Gemini 3,      36)
      (Grok-4,         4)
      (Grok-4.3,      25)
      (Grok-4-fast,   19)
      (EvoCUA-32B,    18)
      (Fara-7B,       17)
    };
\addplot[fill=cAOI!70, draw=cAOI!90]
    coordinates {
      (Claude 4.6,    82)
      (GPT-5.4,       57)
      (Gemini 2.5,    69)
      (Gemini 3,      45)
      (Grok-4,        25)
      (Grok-4.3,      65)
      (Grok-4-fast,   47)
      (EvoCUA-32B,    55)
      (Fara-7B,       34)
    };
\legend{Standard, AOI (full)}

\node[font=\tiny\bfseries, text=cDelta!60!black, fill=white, inner sep=0.8pt, rounded corners=1pt]
    at (axis cs:Claude 4.6, 100) {$\Delta$+44};
\node[font=\tiny\bfseries, text=cDelta!60!black, fill=white, inner sep=0.8pt, rounded corners=1pt]
    at (axis cs:GPT-5.4, 75) {$\Delta$+20};
\node[font=\tiny\bfseries, text=cDelta!60!black, fill=white, inner sep=0.8pt, rounded corners=1pt]
    at (axis cs:Gemini 2.5, 87) {$\Delta$+48};
\node[font=\tiny\bfseries, text=cDelta!60!black, fill=white, inner sep=0.8pt, rounded corners=1pt]
    at (axis cs:Gemini 3, 63) {$\Delta$+9};
\node[font=\tiny\bfseries, text=cDelta!60!black, fill=white, inner sep=0.8pt, rounded corners=1pt]
    at (axis cs:Grok-4, 43) {$\Delta$+21};
\node[font=\tiny\bfseries, text=cDelta!60!black, fill=white, inner sep=0.8pt, rounded corners=1pt]
    at (axis cs:Grok-4.3, 83) {$\Delta$+40};
\node[font=\tiny\bfseries, text=cDelta!60!black, fill=white, inner sep=0.8pt, rounded corners=1pt]
    at (axis cs:Grok-4-fast, 65) {$\Delta$+28};
\node[font=\tiny\bfseries, text=cDelta!60!black, fill=white, inner sep=0.8pt, rounded corners=1pt]
    at (axis cs:EvoCUA-32B, 70) {$\Delta$+37};
\node[font=\tiny\bfseries, text=cDelta!60!black, fill=white, inner sep=0.8pt, rounded corners=1pt]
    at (axis cs:Fara-7B, 52) {$\Delta$+17};

\end{axis}
\end{tikzpicture}
\caption{\textbf{Main results: the AOI lifts every CU model on DynaCU-Bench (100 dynamic
tasks), with zero retraining.} Green numbers are absolute gains in percentage points over
each model's screenshot baseline ($+9$ to $+48$\,pp). Full numbers, confidence intervals,
and the per-model analysis are in Section~\ref{sec:eval}.}
\label{fig:mainresults}
\end{figure}
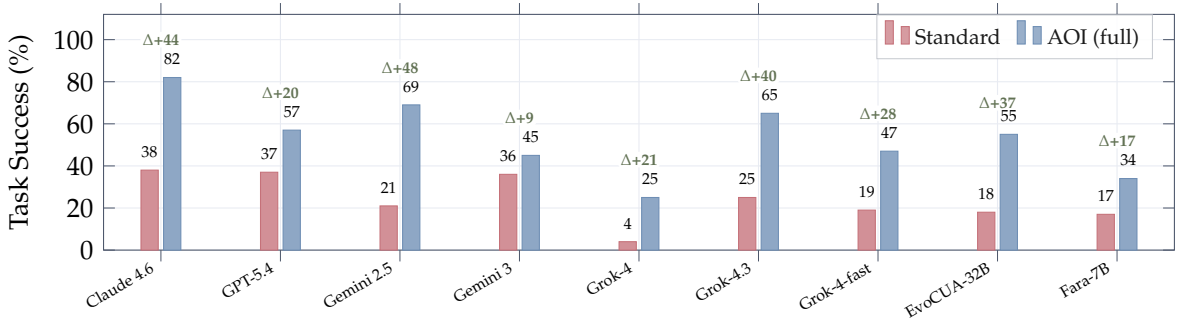

\section{Introduction}
\label{sec:intro}

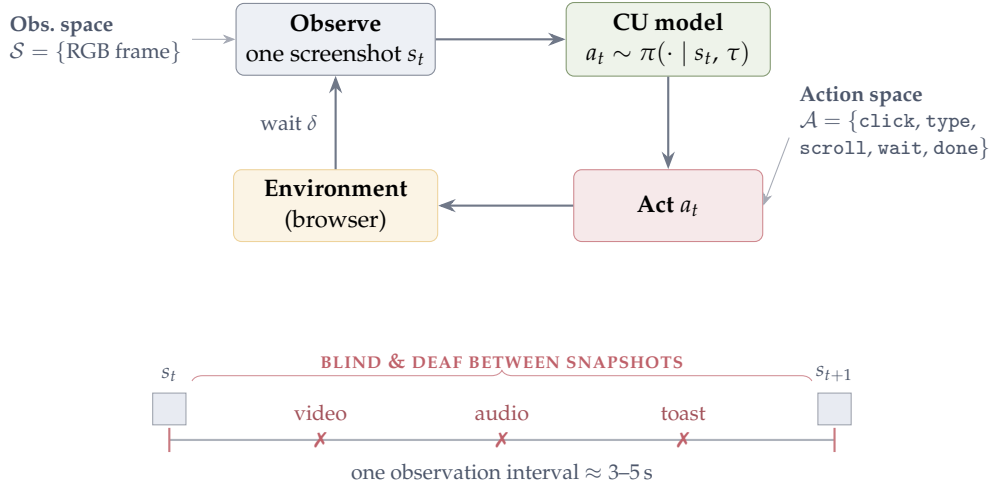
\begin{figure}[t]
\centering
\begin{tikzpicture}[
    >=Stealth,
    font=\small,
    obox/.style={draw, rounded corners=3pt, align=center, font=\footnotesize,
                 minimum height=0.95cm, minimum width=2.5cm},
    sense/.style={obox, fill=cLightGray!45, draw=cGray!70},
    model/.style={obox, fill=cDelta!18, draw=cDelta!80!black, minimum width=2.7cm},
    act/.style={obox, fill=cStandard!14, draw=cStandard!75},
    envb/.style={obox, fill=cAccent!22, draw=cAccent!85, minimum width=2.7cm},
    sp/.style={font=\scriptsize, text=cGray},
  ]

  \node[sense] (obs) at (0,2.2)   {\textbf{Observe}\\one screenshot $s_t$};
  \node[model] (mdl) at (4.4,2.2) {\textbf{CU model}\\$a_t \sim \pi(\cdot \mid s_t,\,\tau)$};
  \node[act]   (act) at (4.4,0)   {\textbf{Act} $a_t$};
  \node[envb]  (env) at (0,0)     {\textbf{Environment}\\(browser)};

  \draw[->, thick, cGray!80] (obs) -- (mdl);
  \draw[->, thick, cGray!80] (mdl) -- (act);
  \draw[->, thick, cGray!80] (act) -- (env);
  \draw[->, thick, cGray!80] (env) -- (obs);
  \node[sp, left=1pt of $(env)!0.5!(obs)$, xshift=-2pt] {wait $\delta$};

  \node[sp, align=left, anchor=east] at (-1.9,2.2) {%
    \textbf{Obs.\ space}\\$\mathcal{S}=\{$RGB frame$\}$};
  \draw[->, cGray!50] (-1.9,2.2) -- (obs.west);

  \node[sp, align=left, anchor=west] at (6.0,1.1) {%
    \textbf{Action space}\\$\mathcal{A}=\{$\texttt{click}, \texttt{type},\\
    \texttt{scroll}, \texttt{wait}, \texttt{done}$\}$};
  \draw[->, cGray!50] (6.0,0.9) -- (act.east);

  \def\ty{-3.1}
  \node[font=\scriptsize\bfseries, text=cStandard] at (2.2,\ty+1.05)
     {\textsc{blind \& deaf between snapshots}};
  \draw[decorate, decoration={brace, amplitude=4pt}, cStandard!70]
     (-1.9,\ty+0.78) -- (6.3,\ty+0.78);
  \draw[thick, cGray!45] (-2.2,\ty) -- (6.6,\ty);
  \foreach \x in {-2.2, 6.6} {
     \draw[thick, cStandard!85] (\x,\ty-0.16) -- (\x,\ty+0.16);
     \draw[cGray!55, fill=cLightGray!60] (\x-0.22,\ty+0.20) rectangle (\x+0.22,\ty+0.62);
  }
  \node[sp, anchor=south] at (-2.2,\ty+0.64) {$s_t$};
  \node[sp, anchor=south] at (6.6,\ty+0.64)  {$s_{t+1}$};
  \node[sp] at (2.2,\ty-0.42) {one observation interval $\approx 3$--$5$\,s};
  \foreach \x/\lab in {-0.2/{video}, 2.2/{audio}, 4.6/{toast}} {
     \node[font=\scriptsize, text=cStandard] at (\x,\ty) {\ding{55}};
     \node[sp, text=cStandard!85!black, anchor=south] at (\x,\ty+0.12) {\lab};
  }

\end{tikzpicture}
\caption{\textbf{The computer-use agent loop and its blind spots.}
A CU agent repeats \emph{observe\,$\to$\,reason\,$\to$\,act}: it captures a
single screenshot $s_t$ (observation space $\mathcal{S}$), the model samples an
action $a_t$ from a small grammar $\mathcal{A}$
(\texttt{click}/\texttt{type}/\texttt{scroll}/\texttt{wait}/\texttt{done}), the
browser executes it, and after a buffer $\delta$ the next screenshot is taken.
The interval is 3--5\,s; \emph{between} snapshots the agent is blind to anything
that moves (video, animation, an auto-dismissing toast) and has no audio channel
at all, so spoken content is never observed.}
\label{fig:culoop}
\end{figure}

\textbf{Computer-use agents are blind between screenshots and deaf to audio.}
A computer-use (CU) agent perceives the screen as a periodic stream of static
screenshots captured every 3--5 seconds and has no audio channel at all
(Figure~\ref{fig:culoop}).
Between snapshots it is \emph{blind}: videos play, slides animate, toast
notifications appear and auto-dismiss, and all of this remains unobserved if the change does not coincide with a screenshot.
Furthermore, it is \emph{deaf}: meeting speeches, spoken instructions, and notification
chimes will never reach the model.
An entire class of everyday computer use is therefore
out of reach, even for agents that handle static interfaces well. This includes watching screencasts, following
meetings, answering voice prompts, and reacting to transient dialogs.

\textbf{The gap is real and unaddressed.}
On static GUI benchmarks such as OSWorld~\citep{osworld}, screenshot agents have
improved from sub-10\% to above 50\% success~\citep{surfer2,uitars2}, approaching the
human baseline in filling forms, navigating sites, editing documents, and operating
professional software.
However, the dynamic-and-audible regime remains a documented but largely unaddressed gap:
video-LLM GUI agents struggle across temporal tasks~\citep{guiworld},
long-context models do \emph{worse} with video input than without~\citep{videowebarena},
paused-frame game evaluations reach only single-digit
completion~\citep{videogamebench}, and screenshot agents miss dozens of hours of temporal dynamics~\citep{cuasuite}. Even a recent CU survey catalogues
many open challenges but not the observation modality itself~\citep{cusurvey}.

\textbf{The observation interface is a separable design axis.}
Why is this hard to fix from the model side?
Retraining CU models on video is expensive, model-specific, and undemonstrated
at scale. Video-native models brute-force frame sampling and waste tokens on
redundant static frames. Meanwhile, native real-time multimodal
APIs~\citep{geminilive,openairealtime} work but lock users to one provider, offer
no selective perception, and require migration (we compare against them in
Section~\ref{sec:streaming}).
We take a different approach, inspired by SWE-agent~\citep{sweagent}. Rather than
training a new model, SWE-agent demonstrated the power of redesigning the agent-computer \emph{action} interface: what commands the model receives and how inputs and feedback are structured.
We argue the \emph{observation} interface is the analogous lever for CU agents.
Prior work focused on enriching \emph{what} a single snapshot encodes, such as accessibility
trees, element overlays, and text surrogates. It left untouched the dimension we target:
\emph{when} the agent looks and through which senses.
Our core principle is \emph{decoupling observation (continuous, adaptive,
multimodal) from action (discrete)}. Every existing CU agent we are aware
of~\citep{anthropic2024claude,openai2025cua,gemini25cu,uitars2,opencua,fara7b,coact}
ties the two together at one static screenshot per step while remaining completely deaf to audio.

\textbf{Our approach.}
The \textbf{Agent-Computer Observation Interface (AOI)} is a lightweight
perception layer placed between the environment and any image-based CU model.
It converts continuous screen and audio streams into the sparse images and text
the model already accepts through three gated components:
inter-step keyframe capture, volume-gated audio transcription, and
model-generated visual narration that persists as text after the source
images are pruned.
On static, silent content the system produces almost no additional input and the behaviour reverts
to the standard loop. On dynamic or spoken content, the model additionally sees what moved
and hears what was said.
No retraining is required, although the optimal component mix is model-specific.

\textbf{Contributions.}
\begin{enumerate}
  \item We identify the \textbf{observation interface} as a separable CU design axis, complementary to SWE-agent's action interface, with \emph{decoupling observation from action} as the core principle (Sections~\ref{sec:intro}--\ref{sec:design}).
  \item We introduce the \textbf{Agent-Computer Observation Interface (AOI)}, a model-agnostic perception layer with three gated components. We release AOI together with \textbf{DynaCU-Bench}, a benchmark of 100 dynamic tasks across 10 categories, and a static task set for degradation control.
  \item Across eight CU models from 7B to frontier scale, the AOI delivers significant gains on dynamic tasks with zero retraining (Section~\ref{sec:eval}), while staying transparent on static work.
  \item We provide a detailed decomposition of the gains: keyframe \emph{selection} is largely irrelevant, most value emerges only once frames are narrated into persistent text, and the component mix must be tuned per model, with one component even becoming harmful on a newer model (Section~\ref{sec:static}).
\end{enumerate}

\section{Background: The CU Agent Loop}
\label{sec:background}

\textbf{The loop.}
Every current CU agent runs the same \emph{observe\,$\to$\,reason\,$\to$\,act}
cycle shown in Figure~\ref{fig:culoop}: it captures a screenshot $s_t$, samples an action
$a_t$ from the model conditioned on $s_t$ and trajectory history $\tau$, executes
the action, waits a short buffer $\delta$, and repeats.
Two spaces define an agent's capabilities.
The \textbf{action space} $\mathcal{A}$ is a small grammar of discrete commands (click, type, scroll,
wait, done) and has been the focus of prior interface work~\citep{sweagent}.
In contrast, the \textbf{observation space} $\mathcal{S}$, the focus of this work, remains limited: today it consists of a single RGB frame, $\mathcal{S}=\{\text{one screenshot}\}$, sampled once per
action.

\textbf{Why one screenshot per step leaves the agent blind.}
The observation interval (model inference, action execution, and
buffer) typically lasts 3--5 seconds, and each screenshot is a single point sample per interval.
Consequently, anything that happens \emph{between} samples is lost: the captured frame is static, rendering
motion invisible, and there is no audio channel so sound is never captured.
Multi-image histories do not help. The extra images are the post-action screenshots
from prior steps and never the moments between them.
Concretely, four categories of content are systematically missed.
(i)~\emph{Transient events} are toasts and dialogs that appear and auto-dismiss
within a single interval.
(ii)~\emph{Continuous visual media} are video, animated tutorials, and transitioning
slides.
(iii)~\emph{Audio} is meeting speech, notification sounds, and spoken prompts.
(iv)~\emph{Periodic noise} is spinners and blinking cursors that should be
\emph{ignored} yet disrupt naive frame differencing methods.
AOI expands $\mathcal{S}$ to cover (i)--(iii) while suppressing (iv), all without
touching the model or $\mathcal{A}$.

\section{System Design}
\label{sec:design}

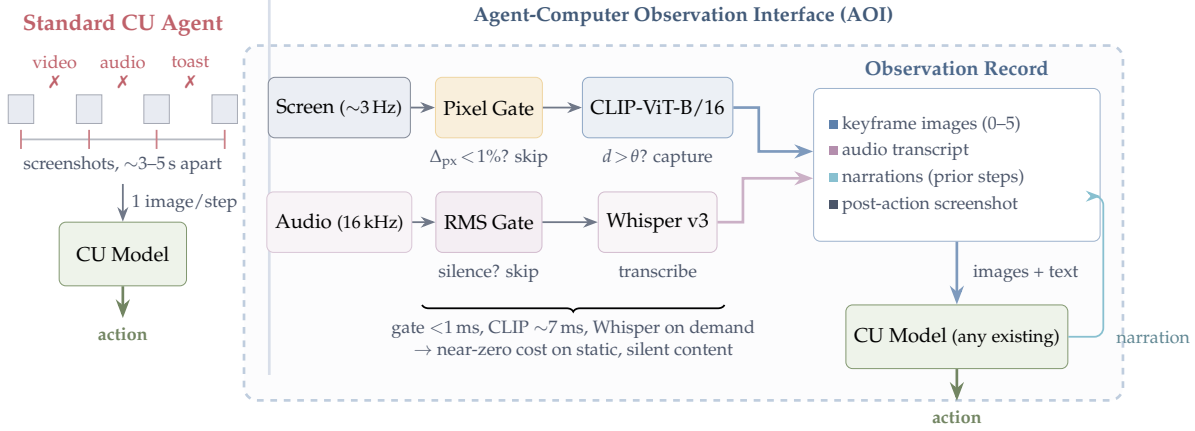
\begin{figure}[t]
\centering
\resizebox{\textwidth}{!}{%
\begin{tikzpicture}[
    >=Stealth,
    box/.style={draw, rounded corners=3pt, minimum height=0.95cm, align=center, font=\footnotesize},
    env/.style={box, fill=cLightGray!50, draw=cGray!70, minimum width=1.6cm},
    gate/.style={box, fill=cAccent!30, draw=cAccent!80, minimum width=1.55cm},
    proc/.style={box, fill=cAOI!12, draw=cAOI!70, minimum width=1.85cm},
    obs/.style={box, fill=white, draw=cAOI!60, minimum width=4.5cm, minimum height=2.4cm},
    mdl/.style={box, fill=cDelta!18, draw=cDelta!80!black, minimum width=2.6cm, minimum height=1cm},
    arr/.style={->, thick, cGray!80},
    harr/.style={->, very thick, cAOI!80},
    lbl/.style={font=\scriptsize, text=cGray},
  ]
  \node[font=\small\bfseries, text=cStandard] at (1.6, 4.2) {Standard CU Agent};

  \draw[thick, cGray!40] (0, 2.4) -- (3.2, 2.4);
  \foreach \x in {0, 1.07, 2.13, 3.2} {
    \draw[thick, cStandard!80] (\x, 2.25) -- (\x, 2.55);
    \draw[cGray!50, fill=cLightGray!60] (\x-0.2, 2.65) rectangle (\x+0.2, 3.1);
  }
  \node[font=\scriptsize, text=cGray] at (1.6, 2.0) {screenshots, $\sim$3--5\,s apart};

  \foreach \x/\t in {0.53/video, 1.6/audio, 2.66/toast} {
    \node[font=\scriptsize, text=cStandard] at (\x, 3.30) {\ding{55}};
    \node[font=\scriptsize, text=cStandard] at (\x, 3.62) {\t};
  }

  \node[mdl, minimum width=2.0cm] (sm) at (1.6, 0.6) {CU Model};
  \draw[arr] (1.6, 1.7) -- (sm.north) node[midway, right, lbl] {1 image/step};
  \draw[arr, cDelta!70!black, very thick] (sm.south) -- ++(0,-0.5)
    node[below, font=\scriptsize\bfseries, text=cDelta!70!black] {action};

  \draw[very thick, cLightGray] (3.9, -1.3) -- (3.9, 4.6);


  \node[env] (scr) at (5.0, 2.9) {Screen {\scriptsize($\sim$3\,Hz)}};
  \node[env, fill=cASR!10, draw=cASR!50] (aud) at (5.0, 1.1) {Audio {\scriptsize(16\,kHz)}};

  \node[gate] (pg) at (7.35, 2.9) {Pixel Gate};
  \node[gate, fill=cASR!15, draw=cASR!60] (rg) at (7.35, 1.1) {RMS Gate};

  \draw[arr] (scr) -- (pg);
  \draw[arr] (aud) -- (rg);

  \node[font=\scriptsize, text=cGray, anchor=north, yshift=-1pt] at (pg.south) {$\Delta_\text{px}\!<\!1\%$? skip};
  \node[font=\scriptsize, text=cGray, anchor=north, yshift=-1pt] at (rg.south) {silence? skip};

  \node[proc] (clip) at (10.0, 2.9) {CLIP-ViT-B/16};
  \node[proc, fill=cASR!8, draw=cASR!60] (wh) at (10.0, 1.1) {Whisper v3};

  \draw[arr] (pg) -- (clip);
  \draw[arr] (rg) -- (wh);

  \node[font=\scriptsize, text=cGray, anchor=north, yshift=-1pt] at (clip.south) {$d\!>\!\theta$? capture};
  \node[font=\scriptsize, text=cGray, anchor=north, yshift=-1pt] at (wh.south) {transcribe};

  \node[obs] (orec) at (14.7, 2.0) {};
  \node[font=\footnotesize\bfseries, text=cAOI!80!black, anchor=south, yshift=2pt] at (orec.north) {Observation Record};
  \node[font=\scriptsize, text=cGray, align=left, anchor=west] at (12.55, 2.00) {%
    \textcolor{cAOI}{\rule{4pt}{4pt}} keyframe images (0--5)\\[2pt]
    \textcolor{cASR}{\rule{4pt}{4pt}} audio transcript\\[2pt]
    \textcolor{cNarr}{\rule{4pt}{4pt}} narrations (prior steps)\\[2pt]
    \textcolor{cGray}{\rule{4pt}{4pt}} post-action screenshot};

  \draw[harr] (clip.east) -- ++(0.4,0) |- ([yshift=6pt]orec.west);
  \draw[harr, cASR!70] (wh.east) -- ++(0.4,0) |- ([yshift=-6pt]orec.west);

  \node[mdl] (cm) at (14.7, -0.7) {CU Model {\scriptsize(any existing)}};
  \draw[harr, very thick] (orec.south) -- (cm.north)
    node[midway, right=3pt, lbl] {images + text};

  \draw[harr, cNarr, thick, rounded corners=4pt]
    (cm.east) -- ++(0.55,0) coordinate (narrcorner) |- ([yshift=-14pt]orec.east);
  \node[font=\scriptsize, text=cNarr!70!black, anchor=west, xshift=2pt] at (narrcorner) {narration};

  \draw[arr, very thick, cDelta!70!black] (cm.south) -- ++(0, -0.5)
    node[below, font=\scriptsize\bfseries, text=cDelta!70!black] {action};

  \draw[thick, decorate, decoration={brace, amplitude=3pt, mirror}]
    (6.30, -0.20) -- (11.05, -0.20);
  \node[font=\scriptsize, text=cGray, align=center, anchor=north] at (8.675, -0.30) {%
    gate $<$1\,ms, CLIP $\sim$7\,ms, Whisper on demand\\
    $\to$ near-zero cost on static, silent content};

  \begin{scope}[on background layer]
    \node[draw=cAOI!40, line width=1.2pt, dashed, rounded corners=6pt, fill=cAOI!2,
          inner xsep=10pt, inner ysep=14pt,
          fit=(scr)(aud)(pg)(rg)(clip)(wh)(orec)(cm),
          label={[font=\footnotesize\bfseries, text=cAOI!70!black, anchor=south, yshift=4pt]above:Agent-Computer Observation Interface (AOI)}] {};
  \end{scope}

\end{tikzpicture}}
\caption{\textbf{Standard CU agents vs.\ AOI-equipped agents.}
\emph{Left:} Current agents observe through periodic screenshots (${\sim}$3--5\,s apart) and are blind and deaf between observations, missing video, audio, and transient UI events (\ding{55}).
\emph{Right:} The AOI continuously monitors screen and audio streams through fast gates ($<$1\,ms) that skip processing for static, silent content.
When gates fire, keyframe extraction (shown here with optional CLIP filtering, see Section~\ref{sec:ablation}) and Whisper ASR produce images and text for the observation record.
Visual narrations feed back as persistent text memory.
The AOI wraps any existing CU model with zero retraining.}
\label{fig:architecture}
\end{figure}

The AOI is inserted between the environment and any existing CU model. It continuously observes the entire interval between agent steps and provides the model with additional keyframes, audio descriptions, and accumulated visual narration, all in the standard image + text format that every CU model already accepts.
On static, silent tasks, the AOI's perception gates remain mostly closed and behaviour reduces to the standard loop wrapped in AOI's structured observation-record format (which does not degrade static work, and Section~\ref{sec:promptperc} shows it slightly helps). On dynamic tasks, the model additionally receives the inter-step keyframes, audio transcripts, and narration.
This design provides strictly more \emph{information}. However, it does not always provide a strictly better \emph{outcome}: the per-step observation work adds latency, which can outweigh potential gains on action-bound tasks (browser games, Section~\ref{sec:eval}). The AOI targets human-paced computer use, not real-time interaction.
No model retraining is needed.

\begin{figure}[t]
\centering
\begin{tcolorbox}[
  colback=gray!3, colframe=black!40,
  title={\small\bfseries Observation Record --- Step $N$ (Meeting task)},
  fonttitle=\small, coltitle=white, colbacktitle=black!60,
  width=0.95\textwidth, boxrule=0.4pt,
]
\ttfamily\scriptsize
\textcolor{blue!70!black}{\# CONTEXT --- text from prior steps (no images)}\\[2pt]
\textbf{Step N-1} (t $\approx$ 18.5--22.0\,s):\\
\quad AUDIO: ``Here's the team. As you can see we\\
\qquad have strong cross-functional coverage.''\\
\quad VISUAL: ``Meeting slide shows Project Team\\
\qquad with 6 members listed in a grid.''\\
\quad ACTION: wait()\\[4pt]
\textcolor{blue!70!black}{\# NEW --- current interval observations}\\[2pt]
\textbf{Step N} (t $\approx$ 22.0--25.5\,s):\\
\quad AUDIO: ``And I want to mention, we've moved\\
\qquad the launch date to April 28th.\\
\qquad Please update your calendars.''\\[2pt]
\quad \textcolor{gray}{[IMAGE: post-action screenshot]}\\[4pt]
\textcolor{blue!70!black}{\# TASK}\\
\quad Read the meeting slides and listen to the\\
\quad discussion. What is the new launch date?\\
\quad Enter it in the Launch Date field.
\end{tcolorbox}
\caption{Example observation record sent to the CU model at step $N$ of a meeting task.
The \textbf{context} section provides text from prior steps (audio transcriptions and visual narrations persist after images are pruned).
The \textbf{new} section contains the current audio transcription and the post-action screenshot (image).
In this case, no keyframes were captured (the slide did not change), so only the screenshot is included as an image.
The task instruction is appended at the end.}
\label{fig:obsrecord}
\end{figure}

\subsection{Architecture Overview}

The AOI consists of three components, each preceded by a fast (sub-millisecond) gate that short-circuits processing when there is nothing new to perceive. Figure~\ref{fig:architecture} contrasts the resulting pipeline with the standard CU loop.
For frames that pass the gates, the additional cost is dominated by CLIP-ViT-B/16 (${\sim}7$\,ms on GPU) for visual frames and Whisper large-v3 (variable, on demand) for audio segments. For the static, silent frames that dominate most workloads, the per-frame cost reduces to the sub-millisecond gate alone.

\begin{enumerate}[label=(\arabic*)]
  \item \textbf{Inter-step keyframe capture}: Continuous screen sampling at ${\sim}3$\,Hz with gating to suppress redundant frames. Produces 0--5 keyframe images per step.
  \item \textbf{Volume-gated audio observer}: RMS energy gate followed by ASR (Whisper large-v3). Produces a text transcription per step, only when audio is present.
  \item \textbf{Visual narration context}: The CU model itself generates a brief text description of new visual information as a side-output of each inference call. These narrations accumulate in the trajectory and persist after keyframe images are pruned from context.
\end{enumerate}

\subsection{Inter-Step Keyframe Capture}
\label{sec:keyframes}

Between agent steps, the screen may change (a video plays, a dialog appears, a slide transitions) or stay static.
The AOI continuously samples the screen at ${\sim}3$\,Hz and selects a subset of frames to include as keyframe images in the observation record (up to 5 per step).
Multiple selection strategies are possible: uniform fixed-rate sampling, pixel-level differencing, random sampling, or semantic filtering via CLIP embeddings~\citep{clip}.
We evaluate five variants spanning these four families (uniform at 1 and 3\,FPS, pixel-diff, random, CLIP) in Section~\ref{sec:ablation} and find that all converge to similar accuracy, so the default implementation uses simple pixel-change gating: if fewer than 1\% of pixels changed since the last captured frame, the sample is skipped (cost: ${<}1$\,ms on CPU).
For static screens, no keyframes are produced and the component has zero overhead.
Algorithm~\ref{alg:keyframe} formalises the two-stage CLIP variant. The default selector drops the CLIP stage (lines~5--10) and uses only Stage~1.

\begin{algorithm}[t]
\caption{Two-stage adaptive keyframe extraction (CLIP variant).
This algorithm is shown for completeness. The selection-method ablation in
Section~\ref{sec:ablation} finds it statistically indistinguishable from
the simpler pixel-only gate (Stage~1) and uniform / random sampling, so the
default selector in every main result is Stage~1 alone.}
\label{alg:keyframe}
\begin{algorithmic}[1]
\REQUIRE Screen sample $x_t$, anchor embedding $e_\text{anchor}$, pixel threshold $\alpha$, CLIP threshold $\theta$
\STATE $\Delta_\text{px} \leftarrow \text{PixelChangeRatio}(x_t, x_\text{prev})$
\IF{$\Delta_\text{px} < \alpha$}
  \RETURN \textsc{Skip} \COMMENT{Stage 1: no pixel change}
\ENDIF
\STATE $e_t \leftarrow \text{CLIP}(x_t)$
\STATE $d \leftarrow 1 - \cos(e_t, e_\text{anchor})$
\IF{$d > \theta$}
  \STATE $e_\text{anchor} \leftarrow e_t$ \COMMENT{Re-anchor}
  \RETURN \textsc{CaptureKeyframe}($x_t$, $t$)
\ENDIF
\RETURN \textsc{Skip} \COMMENT{Stage 2: below semantic threshold}
\end{algorithmic}
\end{algorithm}

\subsection{Volume-Gated Audio Observation}
\label{sec:audio}

Current CU agents have zero audio perception.
They cannot hear meeting speech, notification sounds, or error alerts.
Pure ASR models (Whisper) transcribe speech but produce nothing for non-speech audio events.

\textbf{Volume gate.}
In most computer use, such as file editing, form filling, and silent web browsing, there is no audio.
Computing RMS energy of the audio buffer takes sub-millisecond time.
If the energy falls below a silence threshold, the full ASR call is skipped entirely.
This yields ${\sim}0$\,ms additional cost for the majority of steps.

When audio is present, we transcribe speech with Whisper large-v3~\citep{whisper}, operating on 16\,kHz mono audio captured from the system audio output via PulseAudio virtual devices.
Our implementation focuses on transcribing \emph{speech} only. Non-speech audio events (notification chimes, error beeps) may trip the volume gate but will produce no useful transcript. Consequently, the audio channel's contribution throughout this paper is spoken content only. A multimodal audio model would be needed to interpret the full audio scene (Section~\ref{sec:limitations}).
The audio buffer spans the full inter-step interval, including the post-action buffer. This ensures speech that begins just after an action (e.g.\ a spoken confirmation) is still captured.

\textbf{Overlapping windows.}
Agent step boundaries are determined by model inference latency rather by speech content.
As a result, a typical 3.5-second audio chunk, which captures ${\sim}8$--9 words at normal speaking rate, will almost always fall mid-sentence.
To resolve these boundary artifacts, the audio buffer includes ${\sim}3.5$ seconds of overlap from the previous interval to provide acoustic continuity across step boundaries.

\subsection{Visual Narration for Long-Term Context}
\label{sec:narration}

CU models retain only the most recent images in context, pruning earlier keyframes.
Consequently, on tasks that span many steps, the agent loses all visual memory of prior content.
While audio transcriptions persist naturally as text in the trajectory, visual observations have no analogous persistence.

To address this, at each step the CU model outputs both an action and a brief visual narration, a text description of what is visually new in the current keyframes.
This narration is generated in the same inference call that produces the action, adding no extra inference call and only a small number of output tokens. It persists indefinitely in the trajectory, even after the corresponding images are pruned.
Following the caption-then-reason principle of LLoVi~\citep{llovi}, narrations are generated by the CU model itself (no separate captioner) and are task-relevant rather than generic.

\subsection{Observation Record}
\label{sec:obsrecord}

\begin{figure*}[t]
\centering
\setlength{\tabcolsep}{2pt}
\begin{tabular}{ccc}
\includegraphics[width=0.305\textwidth]{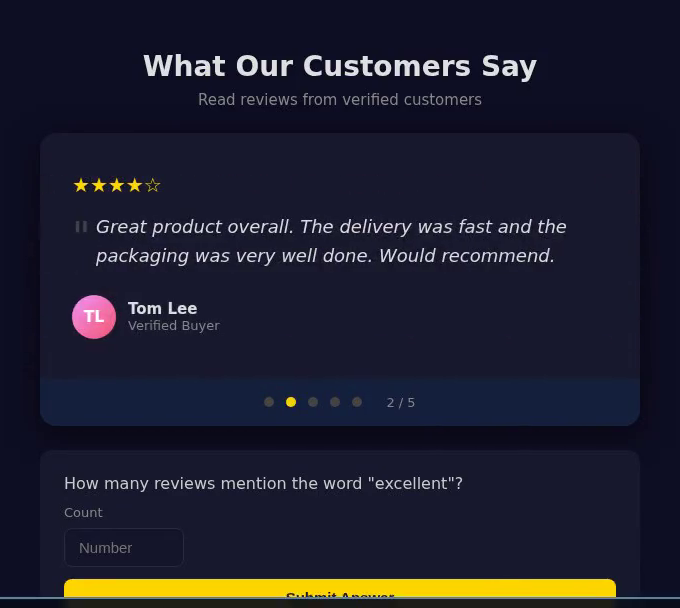} &
\includegraphics[width=0.305\textwidth]{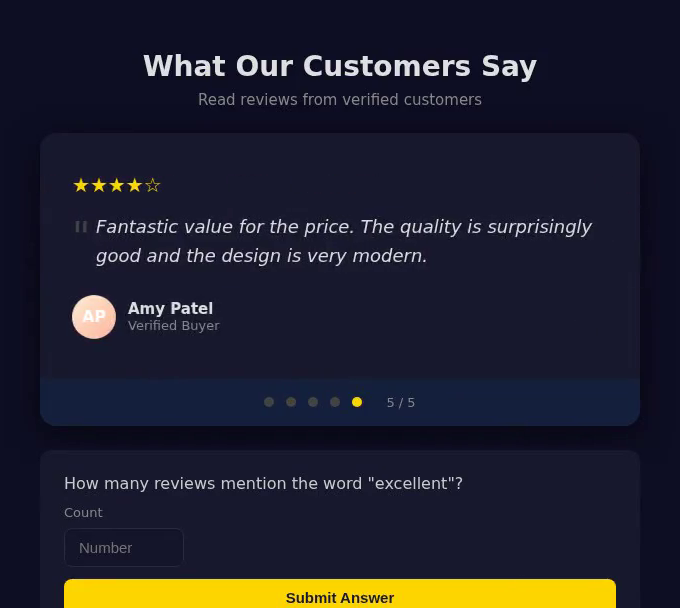} &
\includegraphics[width=0.305\textwidth]{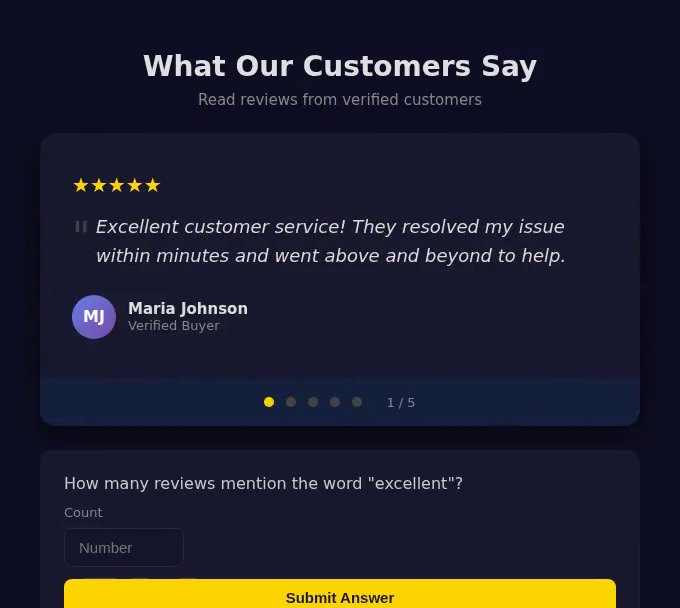} \\
{\footnotesize step 1: testimonial 2/5 (Tom Lee)} &
{\footnotesize step 2: testimonial 5/5 (Amy Patel)} &
{\footnotesize step 3: testimonial 1/5 (Maria Johnson)} \\
\end{tabular}

\vspace{4pt}
\begin{tcolorbox}[colback=gray!3, colframe=black!40, width=0.96\textwidth, boxrule=0.4pt,
  title={\small\bfseries Visual narrations accumulated in the trajectory (verbatim from the evaluation log)},
  fonttitle=\small, coltitle=white, colbacktitle=black!60]
\ttfamily\scriptsize
\textbf{Step 1} (0 keyframes): ``The carousel shows testimonial 2/5 from Tom Lee with a 4-star rating about fast delivery and good packaging, with no mention of `excellent'.''\hfill ACTION: click(567, 400)\\[3pt]
\textbf{Step 2} (1 keyframe): ``The carousel now shows testimonial 5/5 from Amy Patel with a 4-star rating about fantastic value, quality, and modern design --- no mention of `excellent'.''\hfill ACTION: click(567, 401)\\[3pt]
\textbf{Step 3} (2 keyframes): ``The carousel has cycled back to testimonial 1/5 from Maria Johnson with `Excellent customer service!' --- and the transition keyframe also shows a Sarah Robinson testimonial partially visible\ldots''\hfill ACTION: fill(``2'') \textcolor{green!50!black}{\checkmark}
\end{tcolorbox}
\caption{\textbf{What the agent actually sees: real frames and narrations from a recorded benchmark run}
(Carousel task: ``How many reviews mention the word `excellent'?'' Claude Sonnet~4.6 + AOI~full, the run reported in Table~\ref{tab:main}, passed in 3 steps.)
The testimonial carousel rotates every few seconds, so no single screenshot can show more than one of the five reviews.
\emph{Top:} screen states observed across the three agent steps (frames from the run recording, recorder status bar cropped).
\emph{Bottom:} the visual narrations the CU model generated at each step, quoted verbatim from the evaluation log.
The narrations persist as text in the trajectory, letting the agent accumulate evidence across carousel states and answer ``2'' at step~3.}
\label{fig:realkf}
\end{figure*}

At each step, the CU model receives a structured document combining text context from recent prior steps (audio transcriptions, visual narrations, actions taken) and raw observations from the current interval (new audio transcription, any captured keyframe images, and the post-action screenshot).
For static, silent tasks, the raw-observation part of this document reduces to just the post-action screenshot. However, the surrounding structured format (DOM element list, prior-step text trajectory) is still present. As a result, the input is not identical to a raw-screenshot loop. Furthermore, Section~\ref{sec:promptperc} shows this format alone gives a small gain on purely static work.
Figure~\ref{fig:obsrecord} shows a concrete example of the document format and Figure~\ref{fig:realkf} grounds the pipeline in real pixels and verbatim narrations from a recorded benchmark run.

\subsection{Implementation}
\label{sec:implementation}

The AOI is a modular Python layer (${\sim}2{,}600$ lines of code) that wraps any CU model.
Screen and audio are captured through a headless browser and virtual audio
devices. The keyframe encoder runs on the GPU and speech transcription is handled by a separate
CPU service to avoid contention with the model.
Cloud models are called through their native APIs and local models are served on a
single GPU.
Full software, hardware, and hyperparameter details are in
Appendix~\ref{app:implementation}.

\section{DynaCU-Bench}
\label{sec:benchmark}

\subsection{Design Principles}

Existing CU benchmarks~\citep{osworld,webarena,webvoyager} consist almost entirely of tasks solvable from static screenshots.
We introduce \textbf{DynaCU-Bench}, a benchmark of 100 browser-based tasks that specifically require dynamic visual and/or audio perception. These tasks are unsolvable from static screenshots alone.

Design principles:
(i)~Tasks must \emph{require} temporal visual or audio perception.
(ii)~Tasks must represent realistic computer-use scenarios, not artificial constructs.
(iii)~Tasks can be executed in a reproducible headless browser environment.
(iv)~The benchmark covers all four dynamic content categories from Section~\ref{sec:background}.
(v)~The benchmark includes difficulty stratification (3 easy, 4 medium, 3 hard per category).

\subsection{Task Categories}

DynaCU-Bench spans 10 task families across three capability axes: audio
perception (\textsc{Aud}), visual-temporal perception (\textsc{Vis}), and
real-time interaction (\textsc{Int}).
Throughout the paper, we name each family descriptively (Table~\ref{tab:categories}), and use these names consistently in every table and figure.
Each task family has 10 tasks (3 easy, 4 medium, 3 hard). Representative examples
appear in Appendix~\ref{app:tasklist}.

\begin{table}[ht!]
\centering
\caption{DynaCU-Bench task families.
Axes: \textsc{Aud} = audio perception, \textsc{Vis} = visual-temporal
perception, \textsc{Int} = real-time interaction.
Each family contains 3 easy, 4 medium, and 3 hard tasks.}
\label{tab:categories}
\small
\begin{tabular}{@{}llp{7.0cm}@{}}
\toprule
\textbf{Family} & \textbf{Axes} & \textbf{Description} \\
\midrule
\textbf{Podcast}   & \textsc{Aud}             & Listen to spoken narration, then extract facts or answer questions \\
\textbf{Meeting}   & \textsc{Aud+Vis+Int}     & Follow a meeting with slides and speech, then take notes or act \\
\textbf{Screencast}& \textsc{Vis}             & Watch terminal/IDE recordings with auto-advancing frames \\
\textbf{Carousel}  & \textsc{Vis}             & Perceive CSS transitions, rotating content, product carousels \\
\textbf{Dashboard} & \textsc{Vis}             & Monitor live-updating dashboards \\
\textbf{Transient} & \textsc{Vis}             & Respond to toasts and auto-dismissing banners \\
\textbf{Phone}     & \textsc{Aud+Int}         & Listen to a synthesized voice call and respond \\
\textbf{Interview} & \textsc{Aud+Int}         & Answer spoken questions and complete interviews \\
\textbf{Collab}    & \textsc{Vis+Int}         & Handle remote edits in collaborative documents \\
\textbf{Games}     & \textsc{Vis+Int}         & Play interactive games requiring temporal attention \\
\bottomrule
\end{tabular}
\end{table}

\subsection{Evaluation Protocol}

Each task runs in a headless Chromium browser with virtual audio devices for
playback and microphone injection. Spoken content is rendered using
high-quality text-to-speech.
Most tasks (93) are scored by a deterministic check on the resulting page state.
The remaining tasks either combine that check with an LLM quality rubric (6) or rely solely on the LLM judge (1).
A run ends after 15 agent steps or a per-task wall-clock budget (the stimulus
duration plus a fixed margin and any AOI overhead), whichever occurs first.
These budgets bind only for the slowest models
(e.g.\ Grok-4 at $\sim$80--180\,s/call, Section~\ref{sec:grok}).

\section{Evaluation}
\label{sec:eval}

\subsection{Setup}

\textbf{CU models.}
We evaluate six models spanning the capability and scale spectrum.
(i)~Claude Sonnet~4.6 \citep{anthropic2024claude} is closed-source.
(ii)~GPT-5.4 \citep{openai2026gpt54} is closed-source, the latest GPT-5 series tool-calling-capable vision model exposed by OpenAI's chat-completions API at evaluation time.
(iii)~Gemini~2.5~Flash \citep{gemini25cu} is closed-source.
(iv)~Grok-4 \citep{xai2026grok4} is closed-source (xAI), reported in Section~\ref{sec:grok} with attention to the inference-latency confound.
(v)~EvoCUA-32B \citep{evocua} is open-source (32B), with weights and inference recipe released through Meituan's GitHub.
(vi)~Fara-7B \citep{fara7b} is open-source (7B).
In addition, we evaluate two further open-source models accessed through OpenRouter: Qwen3-VL-235B-A22B-Instruct and Qwen3-VL-30B-A3B-Instruct~\citep{qwen3vl}. These serve as a replication on the standard vs.\ AOI~full contrast (Table~\ref{tab:b1_oss}).
We also report three closed-source models released after the main runs --- Gemini~3~Flash, Grok-4.3, and a lower-latency Grok-4-fast-reasoning variant --- on the same contrast (lower rows of Table~\ref{tab:main}, analysed in Sections~\ref{sec:grok}--\ref{sec:why_g3}).
We use each model's vendor-recommended decoding settings (temperature, top-$p$) without modification.
All reported results are measured directly in our DynaCU-Bench harness. We do not rely on any vendor-reported scores.

\textbf{Observation configurations.}
The primary contrast is between \textbf{Standard} (one screenshot per step and no audio, the
current paradigm) and \textbf{AOI~full} (inter-step keyframes + audio
transcription + visual narration). We evaluate both modes across every model.
All remaining modes are ablations and diagnostic controls run on Claude Sonnet~4.6
(unless otherwise noted). A glossary of every mode used in the paper is in
Appendix~\ref{app:modes}.

\subsection{Main Results}

\begin{table}[t]
\centering
\caption{DynaCU-Bench results: task success rate (\%) on the 100 dynamic tasks, one row per model.
Brackets show 95\% Wilson confidence intervals. The \colorbox{bestcolor}{shaded} \textbf{AOI (full)}
column is the best configuration for every model. $\Delta$ is the absolute gain over the standard
baseline; $p$-values are paired McNemar's exact mid-$p$ tests.
``Claude 4.6'' abbreviates Claude Sonnet~4.6 in the figure and text.
Grok-4 is the slowest model at inference (${\sim}80$--180\,s/call), which holds its absolute scores
down even when AOI unblocks perception (see Section~\ref{sec:grok}). The Gemini~3 and the two extra
Grok rows are additional models run after the main cohort on the same harness
(Sections~\ref{sec:grok}--\ref{sec:why_g3}).}
\label{tab:main}
\small
\setlength{\tabcolsep}{5pt}
\begin{tabular}{@{}l c >{\columncolor{bestcolor}}c c c@{}}
\toprule
\textbf{Model} & \textbf{Standard} & \textbf{AOI (full)} & \textbf{$\Delta$ (pp)} & \textbf{$p$ (McNemar)} \\
\midrule
\multicolumn{5}{@{}l}{\textit{Closed-source}}\\
Claude Sonnet 4.6     & 38 \scriptsize[29.1, 47.8] & \textbf{82} \scriptsize[73.3, 88.3] & \textcolor{gaincolor}{$+$44} & \scriptsize $1.3{\times}10^{-10}$ \\
GPT-5.4               & 37 \scriptsize[28.2, 46.8] & \textbf{57} \scriptsize[47.2, 66.3] & \textcolor{gaincolor}{$+$20} & \scriptsize $8.8{\times}10^{-5}$ \\
Gemini 2.5 Flash      & 21 \scriptsize[14.2, 30.0] & \textbf{69} \scriptsize[59.4, 77.2] & \textcolor{gaincolor}{$+$48} & \scriptsize $2.9{\times}10^{-12}$ \\
Gemini 3 Flash        & 36 \scriptsize[27.3, 45.8] & \textbf{45} \scriptsize[35.6, 54.8] & \textcolor{gaincolor}{$+$9}  & \scriptsize $0.18$ \\
Grok-4                & 4 \scriptsize[1.6, 9.8]    & \textbf{25} \scriptsize[17.5, 34.4] & \textcolor{gaincolor}{$+$21} & \scriptsize $3.0{\times}10^{-6}$ \\
Grok-4.3              & 25 \scriptsize[17.5, 34.3] & \textbf{65} \scriptsize[55.3, 73.6] & \textcolor{gaincolor}{$+$40} & \scriptsize $8.2{\times}10^{-10}$ \\
Grok-4-fast-reasoning & 19 \scriptsize[12.5, 27.8] & \textbf{47} \scriptsize[37.5, 56.7] & \textcolor{gaincolor}{$+$28} & \scriptsize $4.9{\times}10^{-6}$ \\
\midrule
\multicolumn{5}{@{}l}{\textit{Open-source}}\\
EvoCUA-32B            & 18 \scriptsize[11.7, 26.7] & \textbf{55} \scriptsize[45.2, 64.4] & \textcolor{gaincolor}{$+$37} & \scriptsize $3.0{\times}10^{-9}$ \\
Fara-7B               & 17 \scriptsize[10.9, 25.5] & \textbf{34} \scriptsize[25.5, 43.7] & \textcolor{gaincolor}{$+$17} & \scriptsize $4.9{\times}10^{-4}$ \\
\bottomrule
\end{tabular}
\end{table}

Table~\ref{tab:main} and Figure~\ref{fig:mainresults} present the main results:
AOI delivers large, consistent gains across every model in Table~\ref{tab:main}. Every improvement is
statistically significant by paired McNemar's exact test ($p<10^{-3}$) except Gemini~3~Flash, whose
muted net gain we trace to a cancelling component in Section~\ref{sec:why_g3}.
\emph{Statistical protocol.}
Each configuration is a single trial of 100 binary tasks (no repeats). We report 95\%
Wilson confidence intervals and evaluate differences with paired McNemar exact
mid-$p$ tests on the discordant tasks. This design applies to every per-family and
ablation cell in the paper. Differences within the ${\sim}3$\,pp decoding-noise band
quantified below should be read as noise unless a paired test says otherwise.

The per-model numbers (Table~\ref{tab:main}) carry three points beyond the raw gains.
\emph{Claude Sonnet~4.6} reaches the highest absolute score. Section~\ref{sec:static} decomposes how its components combine.
\emph{GPT-5.4} gains as reliably as Claude but tops out lower, suggesting its CU capability, not perception, is the binding constraint once observation is adequate.
\emph{Gemini 2.5 Flash} posts the largest absolute gain despite being video-native: the standard agent loop feeds it only periodic screenshots, so its streaming architecture lies idle. Section~\ref{sec:promptperc} (Table~\ref{tab:structured}) splits this into $+25$\,pp from the structured prompt format and $+23$\,pp from new inter-step perception. It benefits from \emph{both}.

Among the open-source models, \emph{EvoCUA-32B} (32B) barely functions on dynamic tasks without the AOI yet rises to rival GPT-5.4's AOI score, while \emph{Fara-7B} (7B), the most reasoning-constrained, still benefits substantially. The pattern that gains track reasoning capacity recurs in the Qwen3-VL replication below.

Two side-by-side trajectories illustrate the mechanism (Figure~\ref{fig:trajectories}, Appendix~\ref{app:perstep}). On a meeting task the standard agent exhausts its entire step budget because it cannot hear the spoken answer, while the AOI agent transcribes it and acts in two steps. On a screencast task the AOI agent's narration captures the answer in a single step.

\textbf{Additional open-source replication (Qwen3-VL family).}
Two more open-source models run through the same contrast via OpenRouter both land inside
the +17 to +48\,pp band of Table~\ref{tab:main} (full numbers in Appendix~\ref{app:fullresults},
Table~\ref{tab:b1_oss}): Qwen3-VL-235B-A22B-Instruct~\citep{qwen3vl} improves +42\,pp
($22 \to 64$, surpassing GPT-5.4's AOI score of 57) and Qwen3-VL-30B-A3B-Instruct improves
+24\,pp ($18 \to 42$). Within the family the gain grows with scale, consistent with the
reasoning-capacity bottleneck seen on Fara-7B.

\textbf{Replication across model generations.}
The lower rows of Table~\ref{tab:main} report three further models, released after the main
runs and evaluated with the same harness, that confirm the gains are not an artifact of one
model cohort: Grok-4.3 improves $+40$\,pp ($25 \to 65$), a lower-latency Grok variant $+28$\,pp
($19 \to 47$), and Gemini~3~Flash $+9$\,pp ($36 \to 45$, $p=0.18$). The two Grok rows control
for the inference-latency floor that holds Grok-4's absolute scores down, and the muted Gemini~3
figure is a cancellation of opposing component effects; both are dissected per-model in
Sections~\ref{sec:grok} and~\ref{sec:why_g3}.

\textbf{Variance.}
Re-running the headline configuration (Claude~+~AOI~full) across two more seeds yields
82\%, 78\%, 76\% (mean 78.7\%, std 3.1\,pp), so the reported 82\% sits at the upper end
of the $\sim$3\,pp decoding-noise band. This spread is orthogonal to the McNemar tests,
which compare two configurations on the \emph{same} stochastic runs rather than measuring
a single score's variability.

\subsection{Per-Family Analysis}

\begin{figure}[t]
\centering
\resizebox{\columnwidth}{!}{%
\begin{tikzpicture}
\begin{axis}[
    width=\columnwidth, height=5.4cm,
    enlargelimits=false, axis on top,
    colormap={div}{color=(cStandard) color=(cStandard!15) color=(white) color=(cDelta!70) color=(gaincolor!60)},
    point meta min=-9, point meta max=9,
    xtick={0,1,2,3,4,5,6,7,8,9},
    xticklabels={Podcast,Meeting,Phone,Interview,Screencast,Carousel,Dashboard,Transient,Collab,Games},
    xticklabel style={rotate=40, anchor=east, font=\scriptsize},
    ytick={0,1,2,3,4},
    yticklabels={Fara-7B,EvoCUA-32B,GPT-5.4,Gemini 2.5,Claude 4.6},
    yticklabel style={font=\scriptsize},
    xtick style={draw=none}, ytick style={draw=none},
    nodes near coords={\pgfmathprintnumber[print sign]\pgfplotspointmeta},
    nodes near coords style={font=\tiny\bfseries, anchor=center, text=black},
    colorbar, colorbar style={width=7pt, font=\scriptsize,
       ytick={-8,-4,0,4,8}, yticklabel={\pgfmathprintnumber[print sign]\tick}},
]
\addplot[matrix plot*, mesh/cols=10, point meta=explicit] coordinates {
 (0,0)[3] (1,0)[4] (2,0)[3] (3,0)[2] (4,0)[2] (5,0)[2] (6,0)[0] (7,0)[-1] (8,0)[2] (9,0)[0]
 (0,1)[8] (1,1)[5] (2,1)[8] (3,1)[2] (4,1)[3] (5,1)[1] (6,1)[2] (7,1)[5] (8,1)[3] (9,1)[0]
 (0,2)[3] (1,2)[1] (2,2)[3] (3,2)[1] (4,2)[3] (5,2)[1] (6,2)[3] (7,2)[6] (8,2)[1] (9,2)[-2]
 (0,3)[8] (1,3)[6] (2,3)[8] (3,3)[2] (4,3)[1] (5,3)[3] (6,3)[8] (7,3)[9] (8,3)[4] (9,3)[-1]
 (0,4)[9] (1,4)[8] (2,4)[8] (3,4)[2] (4,4)[5] (5,4)[5] (6,4)[0] (7,4)[5] (8,4)[5] (9,4)[-3]
};
\end{axis}
\end{tikzpicture}}
\caption{\textbf{Per-family AOI gain} (AOI~full $-$ Standard, in tasks per 10)
across five models, families grouped by primary axis (audio, visual, interaction).
Green = AOI helps, red = AOI hurts.
Gains are positive almost everywhere and largest on the audio families
(Podcast, Phone, Meeting), which are impossible from screenshots alone; the lone
consistent regression is \textsc{Games}, where action latency---not
perception---is the bottleneck. Absolute per-cell scores are in
Appendix~\ref{app:fullresults}.}
\label{fig:heatmap}
\end{figure}
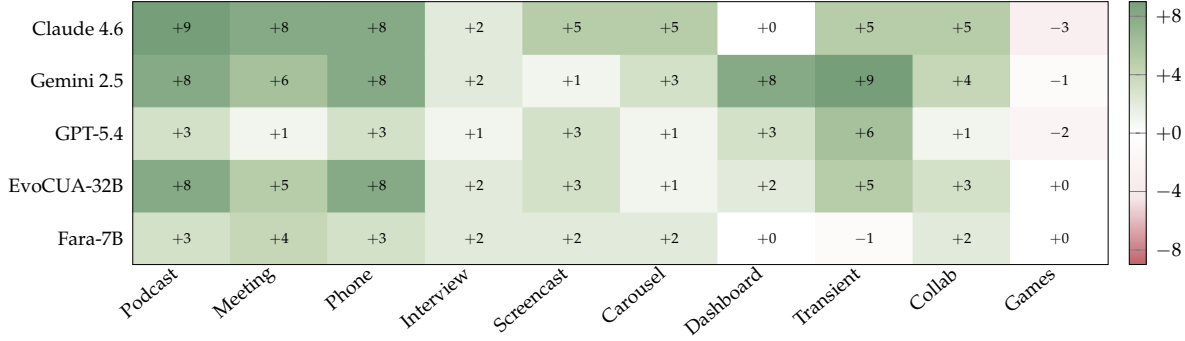

Figure~\ref{fig:heatmap} maps the AOI gain across families and models (full
per-cell scores in Appendix~\ref{app:fullresults}).
Three patterns stand out.
First, the \emph{audio} families (Podcast, Phone, Meeting) improve from near-impossible to
near-solved. These tasks cannot be done from screenshots at all, so audio
perception accounts for nearly the entire gain.
Second, the \emph{visual-temporal} families (Screencast, Carousel, Transient) improve
consistently. The Dashboard family benefits the least, as periodic values are often already visible in a single screenshot.
Finally, \textsc{Games} is the one family that regresses across models. Being tightly action-latency bound rather than perception-limited, the added per-step observation overhead pushes the agent past its reaction window. AOI is designed for human-paced computer use, not real-time interaction.
Two additional model-level observations emerge: the open-source EvoCUA-32B gains most on audio tasks
(weak visual grounding leaves more headroom there), while Dashboard
and Games cells stay flat-to-negative for every model. The gains concentrate on families with a genuine perceptual bottleneck and are absent where screenshots already suffice.

\subsection{Difficulty Breakdown}

The AOI improves performance across every difficulty tier (full breakdown in
Appendix~\ref{app:detailed}, Table~\ref{tab:difficulty}). For Claude Sonnet~4.6, success rates rise from 57\% to 93\% on easy tasks, 38\%\,$\to$\,85\% on medium tasks, and 20\%\,$\to$\,67\% on hard tasks.
Hard tasks remain the most challenging because they need both strong perception \emph{and} reasoning. 
On the smallest model (Fara-7B), gains on medium/hard tasks stay small, while gains on easy tasks nearly double. This highlights that Fara-7B's performance ceiling is primarily limited by reasoning capacity rather than perception.

\subsection{Efficiency}

Better perception leads to fewer but better-aimed steps, resulting in a counter-intuitive
cost consequence: \emph{the AOI is cheaper than the screenshot baseline}. Despite its per-step observation overhead, token costs drop by 15--50\% across all cloud models
(Claude and Gemini~$-50\%$, GPT-5.4~$-15\%$). The reduction in step count (roughly 50\%) outweighs the richer per-step input.
Wall-clock time shows the opposite trend: it \emph{rises} for fast-inference models, where observation work exceeds the latency saved by fewer steps, and falls only
for the slowest model, where step reduction dominates. AOI thus lowers cost for
batch or cost-sensitive workloads. Latency-sensitive use is a per-model judgement call
(full instrumented breakdown in Appendix~\ref{app:detailed}, Table~\ref{tab:efficiency}).

\section{Decomposing the Gain}
\label{sec:static}

Where does the AOI gain come from?
We address this question in three stages. First, we separate the contributions from AOI's prompt format from genuine perception contributions (Section~\ref{sec:promptperc}). Second, we decompose the perception gains channel by channel through targeted ablations (Sections~\ref{sec:ablation}--\ref{sec:keyframe_context}). Third, we turn to per-model case studies showing that the component mix itself must be tuned per model (Sections~\ref{sec:grok}--\ref{sec:diagmap}).

\subsection{Prompt Format vs.\ Perception}
\label{sec:promptperc}

Before attributing the gains to improved perception, we rule out the most concerning
confounding factor: that the AOI helps only by reformatting the prompt.
Two targeted checks confirm this is not the case.

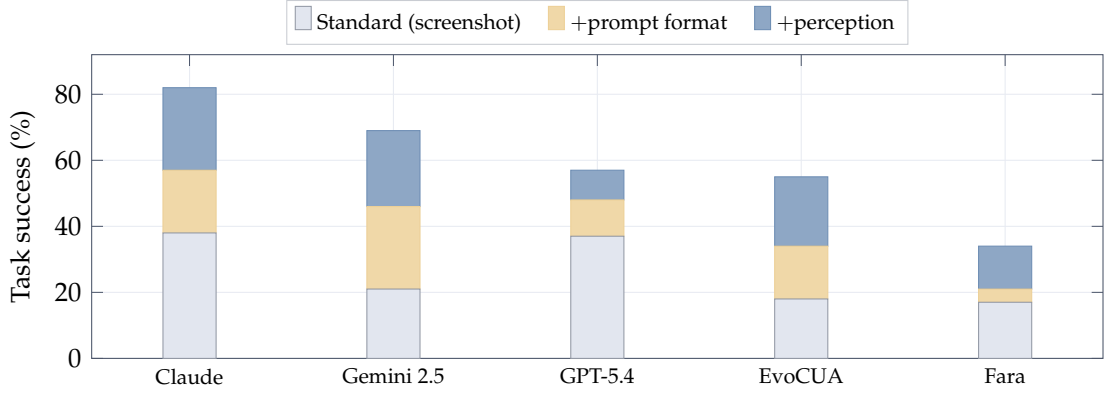
\begin{figure}[t]
\centering
\begin{tikzpicture}
\begin{axis}[
    width=0.95\columnwidth, height=5.6cm, ybar stacked, bar width=20pt,
    ymin=0, ymax=92, ylabel={Task success (\%)}, ylabel style={font=\small},
    symbolic x coords={Claude, Gemini 2.5, GPT-5.4, EvoCUA, Fara},
    xtick=data, xticklabel style={font=\scriptsize},
    ytick={0,20,40,60,80}, yticklabel style={font=\small},
    legend style={at={(0.5,1.18)}, anchor=north, legend columns=3, font=\scriptsize,
                  draw=cGray!30, /tikz/every even column/.append style={column sep=8pt}},
    grid=major, major grid style={cLightGray!60}, axis line style={cGray}, tick style={cGray},
    enlarge x limits=0.12,
]
\addplot[fill=cLightGray!75, draw=cGray!60] coordinates
   {(Claude,38) (Gemini 2.5,21) (GPT-5.4,37) (EvoCUA,18) (Fara,17)};
\addplot[fill=cAccent!75, draw=cAccent!90] coordinates
   {(Claude,19) (Gemini 2.5,25) (GPT-5.4,11) (EvoCUA,16) (Fara,4)};
\addplot[fill=cAOI!70, draw=cAOI!90] coordinates
   {(Claude,25) (Gemini 2.5,23) (GPT-5.4,9) (EvoCUA,21) (Fara,13)};
\legend{Standard (screenshot), $+$prompt format, $+$perception}
\end{axis}
\end{tikzpicture}
\caption{\textbf{The AOI gain splits into a prompt-format and a perception
component---both positive for every model.} Bars stack the raw-screenshot
baseline, the gain from the AOI's structured observation record alone (no
keyframes/audio), and the further gain from perception
(keyframes\,+\,audio\,+\,narration). The worst-case view that the gain is ``just
reformatting'' is ruled out; the split is model-specific (Claude
perception-leaning, Gemini~2.5 prompt-leaning, the small Fara-7B unable to exploit
a richer prompt). Per-model numbers in Appendix~\ref{app:promptdecomp}.}
\label{fig:decomp}
\end{figure}

First, on the \textbf{Static-50} benchmark (50 silent HTML tasks such as forms, tables,
and calculations whose rendered page never changes), the AOI's gates almost
never activate (7 keyframes across all 50 tasks, no audio). Nevertheless, AOI scores a perfect 50/50 compared to the screenshot baseline's 43/50.
This perfect score on a workload with near-zero perception extraction demonstrates a
genuine prompt-format benefit and confirms that AOI does not degrade performance on static tasks (Appendix~\ref{app:promptdecomp}).

Second, on the dynamic benchmark we separate the two components directly. A
control condition that supplies the AOI's structured observation record but disables all
keyframe and audio extraction isolates the prompt-format contribution. The remainder
can be attributed to perception gains.
Figure~\ref{fig:decomp} shows that both components are positive for every model, ruling out both the ``purely reformatting'' and ``purely perception'' interpretations.
The balance is model-specific: Claude is perception-leaning, Gemini~2.5 is
prompt-leaning (for a video-native model, much of the gain is simply a better way
to present context it already had), and the small Fara-7B cannot exploit the
richer prompt at all.
Within the prompt-format component, the DOM-element list is the single largest lever
(Appendix~\ref{app:promptdecomp}).
The rest of this section examines the \emph{perception} component in greater detail: what each
channel (keyframe selection, keyframe images, audio, narration) contributes, and
how.

\subsection{Keyframe Selection Is Immaterial, and Images Pay Off Through Narration}
\label{sec:ablation}

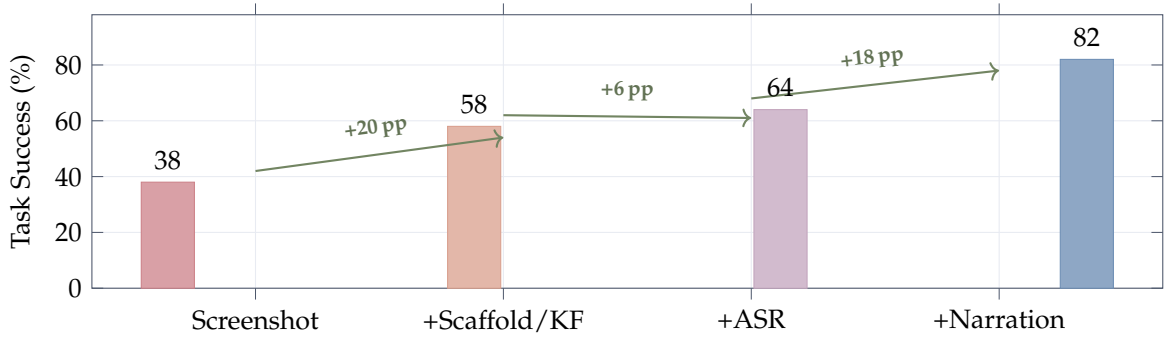
\begin{figure}[t]
\centering
\begin{tikzpicture}
\begin{axis}[
    width=\columnwidth,
    height=5.2cm,
    ybar,
    bar width=20pt,
    ymin=0, ymax=98,
    ylabel={Task Success (\%)},
    ylabel style={font=\small},
    symbolic x coords={Screenshot, {+Keyframes}, {+ASR}, {+Narration}},
    xtick={Screenshot, {+Keyframes}, {+ASR}, {+Narration}},
    xticklabels={Screenshot, {+Scaffold/KF}, +ASR, +Narration},
    xticklabel style={font=\small},
    ytick={0,20,40,60,80},
    yticklabel style={font=\small},
    grid=major,
    major grid style={cLightGray!60},
    axis line style={cGray},
    tick style={cGray},
    enlarge x limits=0.22,
    nodes near coords,
    nodes near coords style={font=\small\bfseries, /pgf/number format/fixed},
    every node near coord/.append style={above=1pt},
]
\addplot[fill=cStandard!60, draw=cStandard!80]
    coordinates {(Screenshot, 38)};
\addplot[fill=cVisual!60, draw=cVisual!80]
    coordinates {({+Keyframes}, 58)};
\addplot[fill=cASR!60, draw=cASR!80]
    coordinates {({+ASR}, 64)};
\addplot[fill=cAOI!70, draw=cAOI!90]
    coordinates {({+Narration}, 82)};

\draw[->, thick, cDelta!70!black]
  (axis cs:Screenshot, 42) -- (axis cs:{+Keyframes}, 54)
  node[midway, above=2pt, font=\scriptsize\bfseries, text=cDelta!60!black, sloped] {+20\,pp};
\draw[->, thick, cDelta!70!black]
  (axis cs:{+Keyframes}, 62) -- (axis cs:{+ASR}, 61)
  node[midway, above=2pt, font=\scriptsize\bfseries, text=cDelta!60!black] {+6\,pp};
\draw[->, thick, cDelta!70!black]
  (axis cs:{+ASR}, 68) -- (axis cs:{+Narration}, 78)
  node[midway, above=2pt, font=\scriptsize\bfseries, text=cDelta!60!black, sloped] {+18\,pp};
\end{axis}
\end{tikzpicture}
\caption{\textbf{Progressive contribution of each AOI component} (Claude Sonnet~4.6).
The first step (+20\,pp) \emph{bundles} inter-step keyframes with the structured
prompt scaffold that bare screenshots lack. The scaffold carries $+19$ of it
(Section~\ref{sec:promptperc}), the keyframe images $+1$\,pp as raw input but
$+10$\,pp once narration is present (Figure~\ref{fig:analysis}b).
ASR adds $+6$\,pp (directional, $p=0.18$). Visual narration adds $+18$\,pp, the
largest single content component. Per-tier significance tests are in
Appendix~\ref{app:stats}.}
\label{fig:ablation}
\end{figure}

Figure~\ref{fig:ablation} walks Claude up the component ladder. Two facts about
the keyframe channel stand out.

\emph{Selection is immaterial.}
Five visual-only selection strategies (uniform 1 and 3\,FPS, pixel-difference,
random, and the two-stage CLIP filter of Algorithm~\ref{alg:keyframe}) all land
in a 56--58\% band and are statistically indistinguishable (every pairwise McNemar
$p>0.5$, Appendix~\ref{app:stats}).
Any inter-step frame breaks the agent's blindness, and \emph{how} it is chosen does
not matter. This result also holds on an open-source model (Qwen3-VL-32B~\citep{qwen3vl},
$p>0.4$ for every pair, Appendix~\ref{app:stats}) and is insensitive to the CLIP
threshold across a $15\times$ range (Appendix~\ref{app:theta}).
No learned or semantic frame selector is needed.

\emph{The images pay off only once narrated.}
The initial $+20$\,pp gain from adding keyframes is driven almost entirely by the structured observation scaffold that bare screenshots lack ($+19$\,pp on its own, Section~\ref{sec:promptperc}). Holding the scaffold fixed, the raw keyframe images themselves contribute only +1\,pp.
However, once the model narrates what it sees, the same keyframes increase their worth to $+10$\,pp
(Figure~\ref{fig:analysis}b). This lift occurs precisely on the task families where
content changes \emph{between} steps (Transient, Carousel), rather than where one
screenshot per step already suffices.
The keyframe channel is thus synergistic with narration. Its model-dependence is
the subject of Section~\ref{sec:keyframe_context}.

\emph{Audio and narration.}
Speech transcription improves performance from 58\% to 64\% (directional at $N{=}100$, $p=0.18$,
and confirmed at scale by the audio-family gains of Figure~\ref{fig:heatmap} and the
Gemini~3 decomposition of Section~\ref{sec:why_g3}), with gains concentrated on the Phone
and Interview families.
Visual narration is the largest single content component, yielding an $+18$\,pp improvement
($64\%\,{\to}\,82\%$, $p=5.3\times10^{-4}$). Whether this benefit arises due to persistent memory
or as inference-time articulation is the question of the next subsection.

\subsection{Narration: Articulation or Persistent Memory?}
\label{sec:narration_discard}

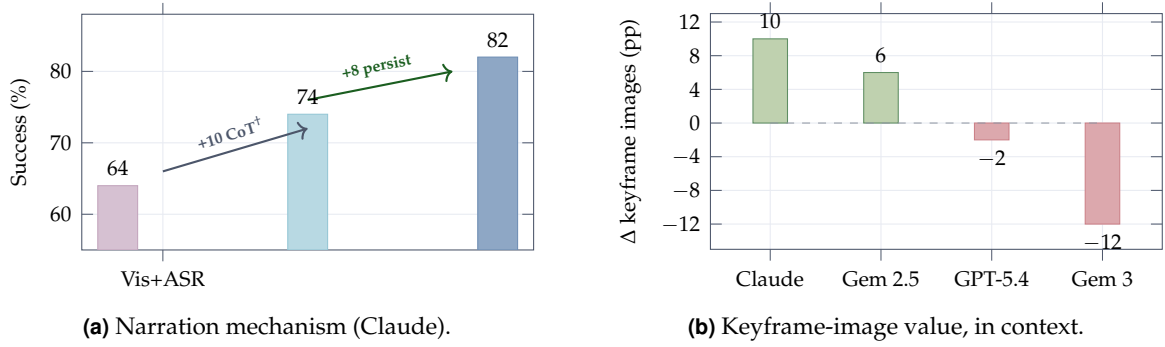
\begin{figure}[t]
\centering
\begin{subfigure}{0.48\textwidth}
\centering
\begin{tikzpicture}
\begin{axis}[
    width=\textwidth, height=4.7cm, ybar, bar width=15pt,
    ymin=55, ymax=88, ylabel={Success (\%)}, ylabel style={font=\scriptsize},
    symbolic x coords={Vis+ASR, Narr.\ disc., Full},
    xtick=data, xticklabel style={font=\scriptsize},
    ytick={60,70,80}, yticklabel style={font=\scriptsize},
    grid=major, major grid style={cLightGray!60},
    axis line style={cGray}, tick style={cGray}, enlarge x limits=0.28,
    nodes near coords, nodes near coords style={font=\scriptsize\bfseries},
]
\addplot[fill=cASR!55,draw=cASR!80] coordinates {(Vis+ASR,64)};
\addplot[fill=cNarr!60,draw=cNarr!90] coordinates {(Narr.\ disc.,74)};
\addplot[fill=cAOI!70,draw=cAOI!90] coordinates {(Full,82)};
\draw[->,thick,cGray] (axis cs:Vis+ASR,66) -- (axis cs:Narr.\ disc.,72)
  node[midway,above,font=\tiny\bfseries,text=cGray,sloped]{+10 CoT$^\dagger$};
\draw[->,thick,gaincolor] (axis cs:Narr.\ disc.,76) -- (axis cs:Full,80)
  node[midway,above,font=\tiny\bfseries,text=gaincolor,sloped]{+8 persist};
\end{axis}
\end{tikzpicture}
\caption{Narration mechanism (Claude).}
\label{fig:narr}
\end{subfigure}\hfill
\begin{subfigure}{0.48\textwidth}
\centering
\begin{tikzpicture}
\begin{axis}[
    width=\textwidth, height=4.7cm, ybar, bar width=13pt,
    ymin=-15, ymax=13, ylabel={$\Delta$ keyframe images (pp)}, ylabel style={font=\scriptsize},
    symbolic x coords={Claude, Gemini 2.5, GPT-5.4, Gemini 3},
    xtick={Claude, Gemini 2.5, GPT-5.4, Gemini 3},
    xticklabels={Claude, Gem 2.5, GPT-5.4, Gem 3},
    xticklabel style={font=\scriptsize},
    ytick={-12,-8,-4,0,4,8,12}, yticklabel style={font=\scriptsize},
    grid=major, major grid style={cLightGray!60},
    axis line style={cGray}, tick style={cGray}, enlarge x limits=0.18,
    nodes near coords, nodes near coords style={font=\scriptsize\bfseries},
    nodes near coords align={vertical},
    every axis plot/.append style={bar shift=0pt},
]
\addplot[draw=gaincolor!70,fill=cDelta!70] coordinates {(Claude,10) (Gemini 2.5,6)};
\addplot[draw=cStandard!80,fill=cStandard!55] coordinates {(GPT-5.4,-2) (Gemini 3,-12)};
\draw[cGray!60,dashed] (axis cs:Claude,0) -- (axis cs:Gemini 3,0);
\end{axis}
\end{tikzpicture}
\caption{Keyframe-image value, in context.}
\label{fig:kfmodel}
\end{subfigure}
\caption{\textbf{What narration and keyframes contribute.}
\textbf{(a)} Visual narration lifts Claude from 64\% (visual+ASR, no narration) to
82\%: discarding the narration text after it is generated keeps 74\%, splitting
the $+18$\,pp into ${\sim}{+}10$\,pp inference-time articulation
($^\dagger$directional, $p{=}0.12$) and $+8$\,pp from persisting it as memory
($p{=}0.039$).
\textbf{(b)} The marginal value of the inter-step keyframe \emph{images}, measured
with audio and narration already present, is model-specific: $+10$/$+6$\,pp on
Claude/Gemini~2.5 (realised only through narration---just $+1$\,pp as raw input),
neutral on GPT-5.4, and a $-12$\,pp \emph{penalty} on Gemini~3 (image-token
dilution, Section~\ref{sec:why_g3}).}
\label{fig:analysis}
\end{figure}

Narration does two things at once: it makes the model \emph{articulate} the new
visual content during reasoning, and it \emph{persists} that content as text that survives
keyframe pruning.
To separate these effects, we run a \emph{narration-discarded} control: the model narrates
exactly as in AOI~full (preserving any inference-time benefit), but the resulting narration
text is dropped before it can be reused.
This condition achieves 74/100, between visual+ASR (64) and full (82)
(Figure~\ref{fig:analysis}a), splitting the $+18$\,pp gain into a directional
$+10$\,pp from articulation ($64\to74$, $p=0.12$, underpowered since $p<0.05$ would
need $N\!\sim\!300$) and a significant $+8$\,pp from persistence
($74\to82$, $p=0.039$).

What is the persistence effect doing?
An external-judge audit of the ten tasks that full AOI wins but narration-discarded
loses finds the prior-step narrations contained the load-bearing fact in
only two cases, both multi-step tasks (breakpoints recorded across video frames
and page titles recorded as a sequence advanced).
In the other eight, the necessary information was present \emph{within} the winning
step itself. In those cases, persistence is not supplying a verbatim answer, but steering
earlier steps onto a better trajectory (three first-step wins are pure decoding
variance).
Narration's primary mechanism is therefore articulation. Persistence matters most
where information genuinely spans steps.

\subsection{The Keyframe Channel Is the Most Model-Dependent}
\label{sec:keyframe_context}

Measured in the deployed configuration (audio and narration already on), the
marginal value of the inter-step keyframe \emph{images} swings from $+10$\,pp on
Claude and $+6$\,pp on Gemini~2.5 (both realised only through narration, since the same
images add $+1$\,pp without it), through a neutral $-2$\,pp on GPT-5.4, to a
$-12$\,pp \emph{penalty} on Gemini~3 (Figure~\ref{fig:analysis}b).
The positive gains concentrates on the families where frames carry novel
between-step content (Transient, Carousel) and is near-zero on audio and static
families, pinning it to narration rather than audio.
The observation generalises into the operative rule of Section~\ref{sec:diagmap}:
information delivered as \emph{text} (transcripts and narration) is robustly
positive on every model, whereas the raw image stream must be treated as a
per-model toggle.
The gates make that toggle cheap: only ${\sim}28\%$ of steps ever produce a
keyframe and ${\sim}14\%$ activate audio, leaving ${\sim}61\%$ of steps fully idle
(per-family activity in Appendix~\ref{app:perstep}).
A cross-engine check (audio re-rendered with a different speech synthesizer and
real terminal-recording screencasts) confirms the audio pipeline does not
silently fail under an unfamiliar acoustic profile (Appendix~\ref{app:realcontent}).

The component balance differs across models not just in degree but in \emph{kind}. Two
models from the replication rows of Table~\ref{tab:main} make this concrete: one whose
binding constraint is inference \emph{latency} rather than perception (Grok), and one
where an AOI component actively \emph{reduces} performance (Gemini~3).

\subsection{Grok: Latency, Not Perception, Is the Ceiling}
\label{sec:grok}
The original Grok-4 scored only 4/100 in the standard setting and 25 with AOI ($+21$\,pp,
$p=3\times10^{-6}$): actions are valid, but a single model call takes 80--180\,s, causing
the agent to time out after one or two steps on harder tasks.
The AOI helps where a few well-placed observations suffice, but cannot unblock
tasks needing many sequential actions, so $+21$\,pp lower-bounds the true effect.
Follow-up evaluations confirm this: a lower-latency variant reaches 47, and the newer
Grok-4.3 reaches 65, decomposing the ceiling into ${\sim}{+}22$\,pp from lowering
latency and ${\sim}{+}18$\,pp from base-model improvement.

\subsection{Gemini~3: A Component Reduces Performance}
\label{sec:why_g3}
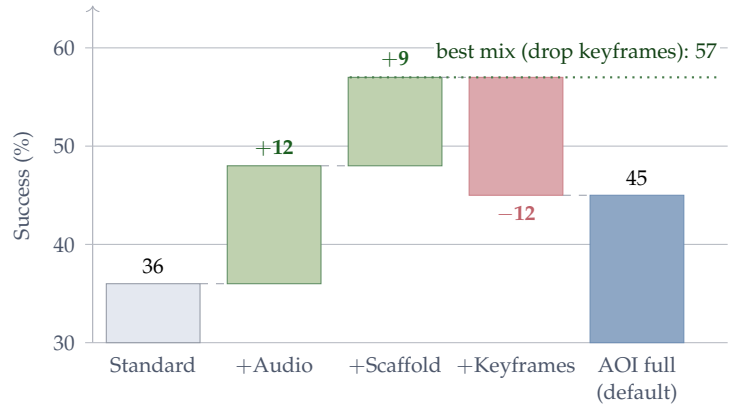
\begin{figure}[t]
\centering
\begin{tikzpicture}[font=\small]
  \def\yb{30}\def\ys{0.13}\def\bw{0.62}
  \draw[cGray!50,->] (0.2,0) -- (0.2,{(62-\yb)*\ys+0.3});
  \foreach \v in {30,40,50,60} {
    \draw[cGray!40] (0.1,{(\v-\yb)*\ys}) -- (8.6,{(\v-\yb)*\ys});
    \node[font=\scriptsize,text=cGray,anchor=east] at (0.1,{(\v-\yb)*\ys}) {\v};
  }
  \node[font=\scriptsize,text=cGray,rotate=90,anchor=south] at (-0.45,{(46-\yb)*\ys}) {Success (\%)};
  \draw[fill=cLightGray!70,draw=cGray!60] (1-\bw,0) rectangle (1+\bw,{(36-\yb)*\ys});
  \node[font=\scriptsize,anchor=south] at (1,{(36-\yb)*\ys}) {36};
  \node[font=\scriptsize,text=cGray,align=center,anchor=north] at (1,-0.05) {Standard};
  \draw[fill=cDelta!70,draw=gaincolor!70] (2.6-\bw,{(36-\yb)*\ys}) rectangle (2.6+\bw,{(48-\yb)*\ys});
  \node[font=\scriptsize\bfseries,text=gaincolor,anchor=south] at (2.6,{(48-\yb)*\ys}) {$+$12};
  \node[font=\scriptsize,text=cGray,align=center,anchor=north] at (2.6,-0.05) {$+$Audio};
  \draw[fill=cDelta!70,draw=gaincolor!70] (4.2-\bw,{(48-\yb)*\ys}) rectangle (4.2+\bw,{(57-\yb)*\ys});
  \node[font=\scriptsize\bfseries,text=gaincolor,anchor=south] at (4.2,{(57-\yb)*\ys}) {$+$9};
  \node[font=\scriptsize,text=cGray,align=center,anchor=north] at (4.2,-0.05) {$+$Scaffold};
  \draw[fill=cStandard!55,draw=cStandard!80] (5.8-\bw,{(45-\yb)*\ys}) rectangle (5.8+\bw,{(57-\yb)*\ys});
  \node[font=\scriptsize\bfseries,text=cStandard,anchor=north] at (5.8,{(45-\yb)*\ys}) {$-$12};
  \node[font=\scriptsize,text=cGray,align=center,anchor=north] at (5.8,-0.05) {$+$Keyframes};
  \draw[fill=cAOI!70,draw=cAOI!90] (7.4-\bw,0) rectangle (7.4+\bw,{(45-\yb)*\ys});
  \node[font=\scriptsize,anchor=south] at (7.4,{(45-\yb)*\ys}) {45};
  \node[font=\scriptsize,text=cGray,align=center,anchor=north] at (7.4,-0.05) {AOI full\\(default)};
  \draw[cGray!55,dashed] (1+\bw,{(36-\yb)*\ys}) -- (2.6-\bw,{(36-\yb)*\ys});
  \draw[cGray!55,dashed] (2.6+\bw,{(48-\yb)*\ys}) -- (4.2-\bw,{(48-\yb)*\ys});
  \draw[cGray!55,dashed] (4.2+\bw,{(57-\yb)*\ys}) -- (5.8-\bw,{(57-\yb)*\ys});
  \draw[cGray!55,dashed] (5.8+\bw,{(45-\yb)*\ys}) -- (7.4-\bw,{(45-\yb)*\ys});
  \draw[gaincolor!80,thick,dotted] (3.6,{(57-\yb)*\ys}) -- (8.5,{(57-\yb)*\ys});
  \node[font=\scriptsize,text=gaincolor!70!black,anchor=east,fill=white,inner sep=1pt]
     at (8.5,{(59.4-\yb)*\ys}) {best mix (drop keyframes): 57};
\end{tikzpicture}
\caption{\textbf{The AOI's small $+9$\,pp net on Gemini~3~Flash is a three-way
cancellation, not saturation.} Audio ($+12$) and the structured scaffold ($+9$)
help as on every other model, but the keyframe-image stream \emph{regresses}
$-12$\,pp---so the default bundle nets only $36\to45$. Dropping just the
keyframes recovers $57/100$ ($+21$\,pp over Standard), showing the slack is real
and the bundle merely mis-tuned. Underlying per-mode and per-family tables are in
Appendix~\ref{app:gemini3}.}
\label{fig:gemini3}
\end{figure}

Gemini~3~Flash~\citep{gemini3flash} nets only $+9$\,pp ($p=0.18$), down from Gemini~2.5's $+48$.
This does not reflect the model having internalised the AOI: a four-way decomposition
(Figure~\ref{fig:gemini3}, with tables in Appendix~\ref{app:gemini3}) shows it is a
\emph{cancellation} of effects.
Audio still helps ($+12$\,pp, with $+10$ on the audio families alone, so Gemini~3 is
not audio-saturated), and the structured scaffold still helps on balance
($+9$\,pp), but the inter-step keyframe images now \emph{regress} $-12$\,pp.
Dropping just the keyframes recovers $57/100$, a $+21$\,pp gain. The slack remains and the default bundle is simply miscalibrated.
A causal probe (Appendix~\ref{app:gemini3}) isolates the mechanism as
\emph{image-token dilution}, where extra images degrade Gemini~3's action grounding
regardless of content, rather than distraction by what the frames show.
The operating rule for this model is simple: deliver observation as text and turn
the keyframe-image stream off.

\subsection{A Diagnostic Map of Model Slack}
\label{sec:diagmap}
Read this way, the AOI's residual gain is a diagnostic of \emph{where} a model has
headroom, along richer axes than perception alone: prompt-handling
(Gemini~2.5 leans prompt-format, $+25/+23$), perception (Claude leans perception,
$+19/+25$), reasoning capacity (Fara-7B cannot exploit a richer prompt,
$+4/+13$), latency (Grok), and action-policy compatibility with the scaffold
(Gemini~3).
The interface is not a fixed bundle. Its components must be selected per model.

\section{Comparison with Streaming Multimodal Baselines}
\label{sec:streaming}

A natural alternative is a \emph{streaming, end-to-end multimodal model} that bundles
perception and reasoning in one trained model: Gemini~Live~\citep{geminilive} and the
OpenAI Realtime API~\citep{openairealtime}. Neither ships as a CU agent; each is a
voice-first assistant with function-calling, which we adapt to DynaCU-Bench (one session
per task, native audio and screenshots streamed over the live websocket, returned tool
calls executed in the browser) and score on the same 12-task audio subset as the AOI.
To test whether a stronger model closes the gap, we also evaluate the current generation:
OpenAI's GA \texttt{gpt-realtime-2}, streaming native audio, in two conditions,
\emph{alone} (screenshot~+~audio) and \emph{+~AOI scaffold} (additionally handed AOI's
\texttt{[PAGE ELEMENTS]} interactive-element list), which separates a perception deficit
from an action-grounding one; and xAI's Grok Voice (\texttt{grok-voice-think-fast}), which
accepts audio and text only and so never sees the screen.

\begin{table}[t]
\centering
\caption{Streaming multimodal baselines vs.\ the AOI on a 12-task audio subset
(3 each from Podcast, Meeting, Phone, Interview), scored on the same task IDs as the
AOI in Table~\ref{tab:main}. ``alone'' is screenshot~+~audio; ``+~scaffold'' adds AOI's
\texttt{[PAGE ELEMENTS]} list. Grok Voice is audio$+$text only (no vision). A
deliberately favourable comparison: no baseline is built for grounded computer use.}
\label{tab:streaming}
\small
\setlength{\tabcolsep}{4.5pt}
\begin{tabular}{@{}l cccc c@{}}
\toprule
\textbf{Configuration} & \textbf{Pod} & \textbf{Meet} & \textbf{Phone} & \textbf{Intv} & \textbf{Total / 12} \\
\midrule
Gemini Live (2.5)                   & 0 & 0 & 0 & 0 & 0 \scriptsize(0\%) \\
OpenAI Realtime (gpt-4o)            & 0 & 0 & 0 & 3 & 3 \scriptsize(25\%) \\
Grok Voice (no vision), audio only  & 1 & 0 & 0 & 0 & 1 \scriptsize(8\%) \\
Grok Voice (no vision), $+$ scaffold & 1 & 0 & 0 & 0 & 1 \scriptsize(8\%) \\
gpt-realtime-2, alone               & 0 & 0 & 1 & 1 & 2 \scriptsize(17\%) \\
gpt-realtime-2, $+$ AOI scaffold    & 2 & 3 & 3 & 3 & 11 \scriptsize(92\%) \\
\rowcolor{bestcolor}
AOI~full (Claude 4.6)               & 3 & 3 & 3 & 3 & \textbf{12} \scriptsize(100\%) \\
\bottomrule
\end{tabular}
\end{table}

A streaming voice model does not substitute for the AOI, and the current generation
sharpens \emph{why} (Table~\ref{tab:streaming}).
The older baselines fail outright: Gemini~Live returns few tool calls and clicks
non-actionable coordinates, and the gpt-4o-backed Realtime succeeds only on the
simplest interview questions.
\texttt{gpt-realtime-2} still fails on its own despite near-perfect transcription
(the logs show it hears every name and figure): the failure is \emph{action grounding},
not perception, as it hears ``Dr.\ Sarah Chen'' yet cannot target the right form field
from a screenshot. Handed AOI's scaffold, the same model recovers most of the gap and
ties the AOI, localising the deficit precisely: a frontier streaming model already
perceives the page but needs AOI's structured grounding to act on it.
Grok Voice makes the same point from the opposite direction: it hears accurately but,
lacking vision, fails with or without the text scaffold, and unlike \texttt{gpt-realtime-2}
cannot exploit the element list, defaulting to blind clicks. Perception without grounded
action is not enough.

\section{Related Work}
\label{sec:related}

\textbf{Expanding the interface, not the model.}
Much agent progress comes from reshaping the agent--computer interface rather than
the weights.
On the \emph{action} side, SWE-agent~\citep{sweagent} redesigned the command set and
feedback format, and agents adopt executable code actions~\citep{codeact} and
element-grounded actions~\citep{mind2web}.
The \emph{observation} side is also well studied, but almost entirely as richer
encodings of a \emph{single static snapshot}: accessibility trees and HTML/DOM text
surrogates~\citep{webarena,mind2web} (which serve as both observation and action
representations), and set-of-mark element overlays that ground vision models on the
screenshot~\citep{setofmark,seeact}.
What none of these representations change is \emph{when} the agent looks: one
snapshot per action, and no audio.
We expand observation along the orthogonal axes they leave out, time (continuous,
between-step capture) and modality (sound). This is complementary to single-frame
encodings and composes with them.

\textbf{Computer-use agents.}
The CU agent landscape spans closed-source systems~\citep{anthropic2024claude,openai2025cua,gemini25cu,surfer2} and open-source models~\citep{uitars2,opencua,fara7b,guiowl,coact}.
Agent~S3~\citep{agents3} demonstrates competitive open-source performance on OSWorld.
A recent CU agent survey~\citep{cusurvey} catalogues many agents and datasets and identifies several open gaps, but does not identify the observation modality as one.
All of these agents operate through the screenshot--reason--act loop.
None addresses dynamic visual content or audio perception.

\textbf{Real-time multimodal models.}
Gemini Live~\citep{geminilive} and the OpenAI Realtime API~\citep{openairealtime} process continuous audio/video/text streams in real time, bundling perception with reasoning into a single trained model.
Our work is complementary: the AOI decouples perception from reasoning, so the same perception layer benefits any CU model without retraining.
Section~\ref{sec:streaming} compares the two paradigms head-to-head on the audio-focused DynaCU-Bench subset.

\textbf{Video understanding for GUI.}
GUI-World~\citep{guiworld}, VideoWebArena~\citep{videowebarena}, and CUA-Suite~\citep{cuasuite} identify the observation problem through benchmarking.
MemGUI-Bench~\citep{memguibench} reveals memory gaps in GUI agents.
These benchmarks identify the problem but do not propose solutions compatible with existing CU agents.

\textbf{Keyframe selection.}
VideoTree~\citep{videotree} and AKS~\citep{aks} use CLIP-based frame selection for offline video QA.
Apollo~\citep{apollo} finds that FPS sampling outperforms uniform sampling and identifies optimal token budgets.
All prior work targets offline video understanding. Our contribution is real-time observation filtering for CU agents, where the two-stage design (sub-millisecond pixel gate followed by ${\sim}7$\,ms CLIP only on changed frames) keeps the per-frame cost low enough to run continuously in the background between agent steps.

\textbf{Adaptive middleware for GUI agents.}
Cradle~\citep{cradle} includes frame difference analysis for games, the closest predecessor to the AOI's perception layer, but uses pixel-level differencing without semantic filtering and has no audio processing.
ScreenLLM~\citep{screenllm} captures stateful screen schemas with keyframe extraction, complementing visual narration but without audio or adaptive observation.
StreamAgent~\citep{streamagent} and Dispider~\citep{dispider} separate perception from response generation for streaming video, paralleling the AOI's architecture, but neither addresses CU agent observation.

\textbf{Structured agent interfaces.}
A complementary line of work bypasses the perception problem by having
websites or applications expose structured interfaces directly to agents:
the Model Context Protocol~\citep{mcp} for tool/data exchange and
declarative web specifications such as NLWeb~\citep{nlweb} for site-side
agent endpoints.
These approaches are powerful where adoption exists, but require website
cooperation. The AOI enables perception on arbitrary, unmodified web
content (any HTML/audio rendered to a browser, including legacy or
third-party pages with no agent endpoint).

\section{Limitations}
\label{sec:limitations}

\textbf{Latency and temporal resolution.}
The AOI extends perception but not decision speed: CU models still take 1--5\,s per
step, and the screen is sampled at ${\sim}3$\,Hz, so sub-${\sim}300$\,ms events and
real-time games fall outside its reach (the Games family regression).
It targets human-paced computer use, not fast interactive control.

\textbf{Perception is not reasoning.}
Better inputs do not improve reasoning over those inputs. On the smallest model the
gain concentrates on easy tasks and reasoning capacity becomes the binding
constraint.

\textbf{Audio is speech-only and narration can err.}
We transcribe speech but not non-speech events (chimes, beeps). A multimodal audio
model would be necessary to cover the full audio scene.
Long-term visual memory also inherits the CU model's mistakes: a misread number
persists in the trajectory.

\textbf{External validity.}
Tasks run in a controlled browser with synthesized audio, so real-world content
(live calls, native apps) may differ, and we do not yet report AOI numbers on
existing dynamic-GUI benchmarks. Those evaluate offline video QA or fixed-context
video rather than a live observe--act loop with a separate audio channel, so
wrapping them with the AOI requires re-instrumenting each harness.
That re-instrumentation is the clearest external-validity test and the main item of
future work.

\textbf{Statistical power.}
Each cell is a single 100-task trial. We use paired McNemar tests for differences,
but the data does not support fine-grained ranking of configurations within a few
points of each other.

\section{Conclusion}
\label{sec:conclusion}

Computer-use agents are blind between screenshots and deaf to audio because
observation is tied to action: one screenshot per step.
We argued the \emph{observation} interface is a separable, underexplored design
axis, the counterpart to the action interface SWE-agent surfaced. We introduced
the AOI, a perception layer that is transparent on static work, adaptive on
dynamic content, and compatible with any existing CU model without retraining.
Across eight models it turns dynamic tasks that were near-impossible from
screenshots into largely solvable ones, while reducing token cost.

Opening up that gain reframes CU perception in three ways. First, keyframe \emph{selection}
is immaterial and value emerges only once frames are narrated into persistent text.
Second, narration helps mainly by making the model articulate what it sees. Third, the
interface is not a fixed bundle. One component even reverses sign on a
later-generation model, so the components must be tuned per model rather than
shipped as one configuration.
We release the perception layer and benchmarks so future work can cleanly separate
observation gains from model-side gains.

\section*{Acknowledgements}

This paper was produced using Pine Copilot's voice-directed \emph{whisper coding} workflow~\citep{pineai2025whispercoding}, in which the authors specify, discuss, and review the work by voice while a coding agent---Claude Code with Claude Opus 4.8---carries out the planning, coding, experiments, and paper writing.
We thank BSQL Networking for hosting the NVIDIA RTX PRO 6000 GPU.

\bibliographystyle{plainnat}
\bibliography{references}

\clearpage

\appendix
\section{Detailed Results}
\label{app:detailed}
\label{app:fullresults}

This appendix provides the per-cell numbers summarised by the figures in the main paper.
\begin{itemize}
  \item Per-family success for every model (Table~\ref{tab:percategory}, the
absolute scores behind the gain heatmap in Figure~\ref{fig:heatmap}).
\item The difficulty breakdown (Table~\ref{tab:difficulty}).
\item Efficiency and cost (Table~\ref{tab:efficiency}).
\item The full per-family ablation sweep (Table~\ref{tab:fullablation}).
\item The additional open-source replication on the Qwen3-VL family (Table~\ref{tab:b1_oss}).
\end{itemize}

\begin{table}[!h]
\centering
\caption{Additional open-source replication on the full 100-task
DynaCU-Bench: Qwen3-VL family via OpenRouter (provider: DeepInfra), evaluated with the same
harness as the models in Table~\ref{tab:main}. $p$-values are
paired McNemar exact tests on the discordant tasks ($b$/$c$ = success
only under standard / only under AOI). Both gains fall inside the
+17 to +48\,pp band of the main table. Two further candidates (UI-TARS-1.5-7B,
GLM-4.5V) were unavailable through our provider list at evaluation time.}
\label{tab:b1_oss}
\small
\begin{tabular}{@{}l cc cc c@{}}
\toprule
\textbf{Model} & \textbf{Standard} & \textbf{AOI full}
& \textbf{$\Delta$ (pp)} & \textbf{$b$/$c$} & \textbf{$p$ (McNemar)} \\
\midrule
Qwen3-VL-235B-A22B-Instruct & 22 & \textbf{64} & $+$42 & 4/46 & $4.5\times10^{-10}$ \\
Qwen3-VL-30B-A3B-Instruct   & 18 & \textbf{42} & $+$24 & 8/32 & $1.8\times10^{-4}$ \\
\bottomrule
\end{tabular}
\end{table}

\begin{table}[!h]
\centering
\caption{Per-family success rates (passes out of 10) for Standard vs.\ AOI~full across the five models
with full per-family breakdowns. Grok-4 is omitted due to latency constraints (see Section~\ref{sec:grok}). These are the absolute scores behind Figure~\ref{fig:heatmap}.}
\label{tab:percategory}
\small
\setlength{\tabcolsep}{2.8pt}
\begin{tabular}{@{}l cc cc cc cc cc@{}}
\toprule
& \multicolumn{2}{c}{\textbf{Claude 4.6}} & \multicolumn{2}{c}{\textbf{GPT-5.4}} & \multicolumn{2}{c}{\textbf{Gemini 2.5}} & \multicolumn{2}{c}{\textbf{EvoCUA-32B}} & \multicolumn{2}{c}{\textbf{Fara-7B}} \\
\cmidrule(lr){2-3}\cmidrule(lr){4-5}\cmidrule(lr){6-7}\cmidrule(lr){8-9}\cmidrule(lr){10-11}
\textbf{Family} & Std & AOI & Std & AOI & Std & AOI & Std & AOI & Std & AOI \\
\midrule
Podcast    & 0 & \textbf{9}  & 0 & 3 & 0 & \textbf{8} & 0 & \textbf{8} & 0 & 3 \\
Meeting    & 2 & \textbf{10} & 2 & 3 & 2 & \textbf{8} & 1 & 6 & 1 & 5 \\
Screencast & 4 & \textbf{9}  & 5 & \textbf{8} & 5 & 6 & 3 & 6 & 3 & 5 \\
Carousel   & 4 & \textbf{9}  & 6 & 7 & 4 & 7 & 3 & 4 & 1 & 3 \\
Dashboard  & 8 & 8           & 6 & \textbf{9} & 1 & \textbf{9} & 2 & 4 & 2 & 2 \\
Transient  & 4 & \textbf{9}  & 2 & \textbf{8} & 0 & \textbf{9} & 0 & 5 & 4 & 3 \\
Phone      & 1 & \textbf{9}  & 1 & 4 & 0 & \textbf{8} & 0 & \textbf{8} & 1 & 4 \\
Interview  & 7 & \textbf{9}  & 6 & 7 & 5 & 7 & 6 & \textbf{8} & 1 & 3 \\
Collab     & 3 & \textbf{8}  & 4 & 5 & 2 & 6 & 2 & 5 & 2 & 4 \\
Games      & 5 & 2           & 5 & 3 & 2 & 1 & 1 & 1 & 2 & 2 \\
\midrule
\textbf{Total} & 38 & \textbf{82} & 37 & \textbf{57} & 21 & \textbf{69} & 18 & \textbf{55} & 17 & \textbf{34} \\
\bottomrule
\end{tabular}
\end{table}

\begin{table}[!h]
\centering
\caption{Task success by difficulty (Standard $\to$ AOI~full).
Easy: 30 tasks. Medium: 40. Hard: 30.}
\label{tab:difficulty}
\small
\begin{tabular}{@{}l ccc ccc@{}}
\toprule
& \multicolumn{3}{c}{\textbf{Standard}} & \multicolumn{3}{c}{\textbf{AOI Full}} \\
\cmidrule(lr){2-4}\cmidrule(lr){5-7}
\textbf{Model} & Easy & Med & Hard & Easy & Med & Hard \\
\midrule
Claude 4.6 & 17/30 & 15/40 & 6/30 & \textbf{28}/30 & \textbf{34}/40 & \textbf{20}/30 \\
GPT-5.4 & 15/30 & 17/40 & 5/30 & 21/30 & 26/40 & 10/30 \\
Gemini 2.5 & 10/30 & 9/40 & 2/30 & 24/30 & 33/40 & 12/30 \\
EvoCUA-32B & 6/30 & 8/40 & 4/30 & 20/30 & 26/40 & 9/30 \\
Fara-7B & 9/30 & 5/40 & 3/30 & 19/30 & 11/40 & 4/30 \\
\bottomrule
\end{tabular}
\end{table}

\begin{table}[!h]
\centering
\caption{Efficiency and cost on the 100-task DynaCU-Bench. Grok-4 is omitted due to inference latency constraints that distort the comparison.
Steps = mean steps per task. Time = mean wall-clock per task. Lat.\ = total inference
latency per task. Obs = total observation processing time per task. Tokens = input + output tokens per task
(estimated, same estimator for every row). \$/100 = cost per 100 tasks at May 2026 list prices
(local models free).
The AOI reduces token cost on every cloud model because the reduction in step count more than offsets
the increased input size.}
\label{tab:efficiency}
\small
\setlength{\tabcolsep}{4.0pt}
\begin{tabular}{@{}l cc cc cc@{}}
\toprule
\textbf{Configuration} & \textbf{Steps} & \textbf{Time (s)} & \textbf{Lat.\ (s)} & \textbf{Obs (s)} & \textbf{Tokens (k)} & \textbf{\$/100} \\
\midrule
\multicolumn{7}{l}{\emph{Claude Sonnet~4.6}} \\
\quad Standard & 10.7 & 40.6 & 26.3 & 2.5 & 7.6 & \$2.72 \\
\quad AOI full & \textbf{4.8} & 46.2 & 15.2 & 12.0 & \textbf{3.8} & \textbf{\$1.35} \\
\midrule
\multicolumn{7}{l}{\emph{GPT-5.4}} \\
\quad Standard & 10.0 & 26.6 & 14.4 & 2.4 & 5.8 & \$3.23 \\
\quad AOI full & \textbf{7.6} & 50.6 & 11.2 & 17.8 & \textbf{5.0} & \textbf{\$2.76} \\
\midrule
\multicolumn{7}{l}{\emph{Gemini 2.5 Flash}} \\
\quad Standard & 11.5 & 55.3 & 40.4 & 2.6 & 7.6 & \$0.32 \\
\quad AOI full & \textbf{5.4} & 61.9 & 20.6 & 20.6 & \textbf{4.0} & \textbf{\$0.16} \\
\midrule
\multicolumn{7}{l}{\emph{EvoCUA-32B (local)}} \\
\quad Standard & 9.3 & 117.0 & 99.0 & 2.3 & 5.8 & --- \\
\quad AOI full & \textbf{5.3} & 113.9 & 72.0 & 22.6 & \textbf{4.0} & --- \\
\midrule
\multicolumn{7}{l}{\emph{Fara-7B (local)}} \\
\quad Standard & 13.4 & 35.1 & 24.0 & 2.5 & 9.2 & --- \\
\quad AOI full & \textbf{9.3} & 90.7 & 17.0 & 31.6 & \textbf{7.1} & --- \\
\bottomrule
\end{tabular}
\end{table}

\begin{table}[!h]
\centering
\caption{Complete per-family results for all 8 observation configurations tested
with Claude Sonnet~4.6. Family abbreviations: Pod=Podcast, Meet=Meeting,
Scr=Screencast, Car=Carousel, Dash=Dashboard, Tran=Transient, Phn=Phone,
Int=Interview, Col=Collab, Gam=Games.}
\label{tab:fullablation}
\small
\setlength{\tabcolsep}{3pt}
\begin{tabular}{@{}l cccccccccc c@{}}
\toprule
\textbf{Mode} & Pod & Meet & Scr & Car & Dash & Tran & Phn & Int & Col & Gam & \textbf{Total} \\
\midrule
Standard & 0 & 2 & 4 & 4 & 8 & 4 & 1 & 7 & 3 & 5 & 38 \\
Uniform 1\,FPS & 0 & 4 & 10 & 8 & 9 & 8 & 1 & 8 & 6 & 4 & 58 \\
Uniform 3\,FPS & 2 & 5 & 8 & 8 & 8 & 8 & 1 & 7 & 6 & 3 & 56 \\
Pixel-diff & 1 & 5 & 9 & 7 & 9 & 8 & 1 & 8 & 6 & 4 & 58 \\
Random keyframes & 1 & 6 & 8 & 6 & 9 & 8 & 1 & 8 & 6 & 3 & 56 \\
AOI visual & 1 & 5 & 9 & 7 & 8 & 9 & 1 & 8 & 6 & 4 & 58 \\
AOI visual+ASR & 1 & 4 & 10 & 9 & 9 & 9 & 4 & 9 & 6 & 3 & 64 \\
\textbf{AOI full} & \textbf{9} & \textbf{10} & \textbf{9} & \textbf{9} & \textbf{8} & \textbf{9} & \textbf{9} & \textbf{9} & \textbf{8} & 2 & \textbf{82} \\
\bottomrule
\end{tabular}
\end{table}

\section{Prompt-Format vs.\ Perception Decomposition}
\label{app:promptdecomp}

These are the numbers behind Figure~\ref{fig:decomp}.
\begin{itemize}
  \item Table~\ref{tab:static}: The Static-50 no-degradation check.
  \item Table~\ref{tab:structured}: The per-model prompt/perception split.
  \item Table~\ref{tab:a2_decomp}: Fine-grained breakdown of the prompt-format effect, showing that a DOM-element
list is the single largest contributor.
\end{itemize}

\begin{table}[!h]
\centering
\caption{Static-50 (50 purely static, silent tasks) on Claude Sonnet~4.6. The gates
suppress almost all extraction (7 keyframes total, no audio), yet the AOI scores
50/50. This shows the presence of a prompt-format effect, not improved perception, as well as no degradation on static work.}
\label{tab:static}
\small
\begin{tabular}{@{}l cc cc@{}}
\toprule
\textbf{Mode} & \textbf{Pass/50} & \textbf{Avg Steps} & \textbf{Total KF} & \textbf{Audio Seg.} \\
\midrule
Standard & 43 \scriptsize(86\%) & 3.26 & 0 & 0 \\
AOI full & \textbf{50} \scriptsize(100\%) & \textbf{1.62} & 7 & 0 \\
\bottomrule
\end{tabular}
\end{table}

\begin{table}[!h]
\centering
\caption{Prompt-format vs.\ perception decomposition on DynaCU-Bench (100 tasks).
A structured-record control with no keyframes/audio isolates the prompt format gain
$\Delta_\text{prompt}$. The remainder up to AOI~full is pure perception gain $\Delta_\text{percept}$.
Both components are positive for every model. (Results for Gemini~3~Flash, where the split differs, is in
Appendix~\ref{app:gemini3}.)}
\label{tab:structured}
\small
\setlength{\tabcolsep}{3.0pt}
\begin{tabular}{@{}l ccc cc@{}}
\toprule
\textbf{Model} & \textbf{Standard} & \textbf{$+$format} & \textbf{AOI full}
& \textbf{$\Delta_\text{prompt}$} & \textbf{$\Delta_\text{percept}$} \\
\midrule
Claude Sonnet 4.6   & 38 & 57 & 82 & $+$19 & $+$25 \\
GPT-5.4             & 37 & 48 & 57 & $+$11 & $+$9 \\
Gemini 2.5 Flash    & 21 & 46 & 69 & $+$25 & $+$23 \\
EvoCUA-32B          & 18 & 34 & 55 & $+$16 & $+$21 \\
Fara-7B             & 17 & 21 & 34 & $+$4  & $+$13 \\
\bottomrule
\end{tabular}
\end{table}

\begin{table}[!h]
\centering
\caption{Which prompt sub-component carries the load (Claude Sonnet~4.6, 100
tasks). Trajectory = prior-step action/text history and sectioning. PAGE EL =
DOM-element list. The element list is the single largest lever ($+30$\,pp added to
the minimal prompt), and the two are sub-additive.}
\label{tab:a2_decomp}
\small
\begin{tabular}{@{}l ccc@{}}
\toprule
\textbf{Mode} & \textbf{Pass/100} & \textbf{Trajectory?} & \textbf{Element list?} \\
\midrule
Minimal (task + screenshot) & 16 & no  & no  \\
$+$trajectory (=\,Standard) & 38 & yes & no  \\
$+$element list             & 46 & no  & yes \\
$+$both (=\,$+$format)      & \textbf{57} & yes & yes \\
\bottomrule
\end{tabular}
\end{table}

\section{Statistical Tests and Component Decompositions}
\label{app:stats}

This appendix includes:
\begin{itemize}
  \item The per-tier significance tests behind Figure~\ref{fig:ablation} (Table~\ref{tab:ablation_stats}).
\item The open-source replication of the selection-invariance finding (Table~\ref{tab:oss_selection}).
\item The per-family narration decomposition behind Figure~\ref{fig:analysis}a (Table~\ref{tab:narration}).
\item The in-context keyframe-image decomposition behind Figure~\ref{fig:analysis}b (Table~\ref{tab:keyframe_context}).
\end{itemize}

\begin{table}[!h]
\centering
\caption{Pairwise McNemar exact mid-$p$ tests between successive ablation tiers
(Claude Sonnet~4.6, 100 tasks). $b$/$c$ are discordant counts (success only in
the previous / only the current configuration). The selection-method group is
statistically indistinguishable. The between-tier transitions are significant.}
\label{tab:ablation_stats}
\small
\begin{tabular}{@{}l rrr l@{}}
\toprule
\textbf{Comparison} & \textbf{Total} & \textbf{$b$} & \textbf{$c$} & \textbf{$p$} \\
\midrule
Standard $\to$ Uniform 1\,FPS         & 38 / 58 & 3 & 23 & $8.8 \times 10^{-5}$ \\
Uniform 1\,FPS $\to$ Uniform 3\,FPS   & 58 / 56 & 7 & 5  & $0.77$ \\
Uniform 3\,FPS $\to$ Pixel-diff       & 56 / 58 & 4 & 6  & $0.75$ \\
Pixel-diff $\to$ Random keyframes      & 58 / 56 & 4 & 2  & $0.69$ \\
Random keyframes $\to$ CLIP adaptive   & 56 / 58 & 4 & 6  & $0.75$ \\
CLIP adaptive $\to$ AOI visual+ASR     & 58 / 64 & 4 & 10 & $0.18$ \\
AOI visual+ASR $\to$ AOI full          & 64 / 82 & 4 & 22 & $5.3 \times 10^{-4}$ \\
\bottomrule
\end{tabular}
\end{table}

\begin{table}[!h]
\centering
\caption{Selection-method ablation replicated on an open-source model
(Qwen3-VL-32B), 50 visual-temporal tasks (Screencast, Carousel, Dashboard,
Transient, Games). The three methods cluster with no pairwise significance
($p>0.4$), mirroring Claude: the convergence is not model-specific.}
\label{tab:oss_selection}
\small
\begin{tabular}{@{}l c c@{}}
\toprule
\textbf{Selection method} & \textbf{Pass / 50} & \textbf{Vs.\ adjacent} \\
\midrule
Uniform 1\,FPS    & 26 & --- \\
Pixel-diff        & 27 & vs.\ Uniform: $p=0.69$ \\
Random keyframes  & 27 & vs.\ Pixel-diff: $p=1.00$ \\
\bottomrule
\end{tabular}
\end{table}

\begin{table}[!h]
\centering
\caption{Narration decomposition (Claude Sonnet~4.6, 100 tasks): visual+ASR (no
narration) $\to$ narration-discarded $\to$ full. The discarded run isolates the
inference-time articulation effect, and the gap to full isolates persistence.
Per-family rows are descriptive (single-trial). Note Transient reads $9{\to}4{\to}9$,
a decoding fluctuation on a 10-task cell that the totals average out.}
\label{tab:narration}
\small
\setlength{\tabcolsep}{3pt}
\begin{tabular}{@{}l c cccccccccc l@{}}
\toprule
\textbf{Configuration} & \textbf{Tot.} & Pod & Meet & Scr & Car & Dash & Tran & Phn & Int & Col & Gam & \textbf{Vs.\ adj.} \\
\midrule
Visual+ASR (no narr.)        & 64 & 1 & 4 & 10 & 9 & 9 & 9 & 4 & 9 & 6 & 3 & --- \\
Full, narration discarded    & 74 & 9 & 10 & 8 & 8 & 7 & 4 & 9 & 9 & 7 & 3 & $p=0.12$ \\
\rowcolor{bestcolor}
Full (narration persisted)   & \textbf{82} & 9 & 10 & 9 & 9 & 8 & \textbf{9} & 9 & 9 & 8 & 2 & $p=0.039$ \\
\bottomrule
\end{tabular}
\end{table}

\begin{table}[!h]
\centering
\caption{Marginal value of the inter-step keyframe \emph{images} measured in
context (audio and narration present): scaffold+ASR+narration without keyframes
vs.\ with them. $\Delta_\text{KF}$ ranges from $+10$\,pp to a $-12$\,pp penalty
across models, the most model-dependent component.
$^\dagger$GPT-5.4 here is measured through a single backend for both modes, so
only its within-row $\Delta_\text{KF}$ is comparable to the absolute scores
elsewhere.}
\label{tab:keyframe_context}
\small
\setlength{\tabcolsep}{6pt}
\begin{tabular}{@{}l ccc@{}}
\toprule
\textbf{Model} & \textbf{no keyframes} & \textbf{$+$keyframes} & \textbf{$\Delta_\text{KF}$ (pp)} \\
\midrule
Claude Sonnet 4.6 & 72 & \textbf{82} & $+$10 \\
Gemini 2.5 Flash  & 63 & \textbf{69} & $+$6 \\
GPT-5.4$^\dagger$ & \textbf{72} & 70 & $-$2 \\
Gemini 3 Flash    & \textbf{57} & 45 & $-$12 \\
\bottomrule
\end{tabular}
\end{table}

\section{Per-Step Observation Activity}
\label{app:perstep}

\begin{figure}[!h]
\centering
\begin{tikzpicture}
\begin{axis}[
    width=0.95\columnwidth, height=5cm, xbar stacked, bar width=10pt,
    xmin=0, xmax=100, xlabel={Percentage of steps (\%)}, xlabel style={font=\small},
    symbolic y coords={Games, Collab, Interview, Phone, Transient, Dashboard, Carousel, Screencast, Meeting, Podcast},
    ytick=data, yticklabel style={font=\small}, xticklabel style={font=\small},
    legend style={at={(0.98,0.03)}, anchor=south east, font=\scriptsize,
                  draw=cGray!30, fill=white, fill opacity=0.92, legend columns=2, column sep=4pt},
    grid=major, major grid style={cLightGray!40}, axis line style={cGray},
    tick style={cGray}, enlarge y limits=0.08,
]
\addplot[fill=cAOI!65, draw=cAOI!80] coordinates
    {(0,Podcast) (23.5,Meeting) (25.6,Screencast) (60.7,Carousel) (28.8,Dashboard) (15.8,Transient) (0,Phone) (0,Interview) (2.6,Collab) (36.4,Games)};
\addplot[fill=cASR!60, draw=cASR!80] coordinates
    {(34.3,Podcast) (47.0,Meeting) (0,Screencast) (0,Carousel) (0,Dashboard) (0,Transient) (26.3,Phone) (37.0,Interview) (10.3,Collab) (0,Games)};
\addplot[fill=cNarr!60, draw=cNarr!80] coordinates
    {(8.6,Podcast) (26.5,Meeting) (2.6,Screencast) (0,Carousel) (0,Dashboard) (0,Transient) (0,Phone) (0,Interview) (7.7,Collab) (0,Games)};
\addplot[fill=cLightGray!50, draw=cLightGray!80] coordinates
    {(57.1,Podcast) (3.0,Meeting) (71.8,Screencast) (39.3,Carousel) (71.2,Dashboard) (84.2,Transient) (73.7,Phone) (63.0,Interview) (79.5,Collab) (63.6,Games)};
\legend{Visual KF, Audio, Both, Idle (gates skip)}
\end{axis}
\end{tikzpicture}
\caption{\textbf{Observation activity per family.} Percentage of steps in which
each channel fires, and gray is idle (both gates suppress). Carousel is highly visual.
Phone and Interview are audio-only. Meeting uses both. ${\sim}61\%$ of steps are
idle overall.}
\label{fig:obsactivity}
\end{figure}
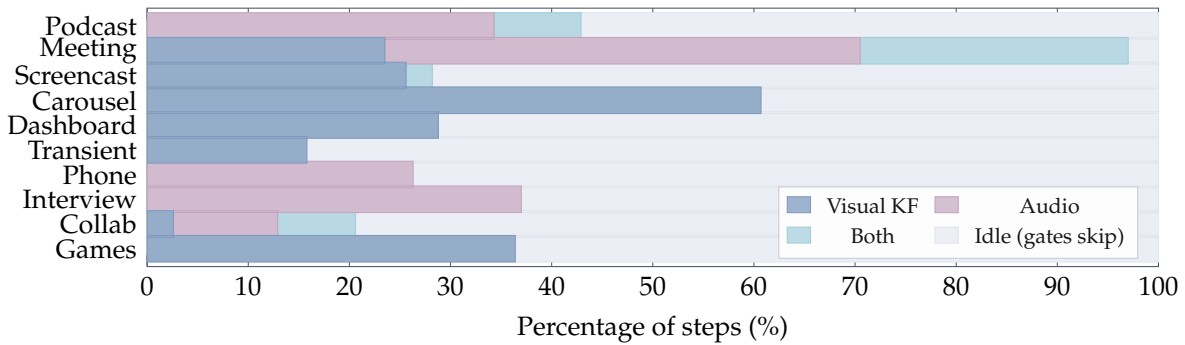

The gates suppress processing on the majority of steps: only ${\sim}28\%$ produce
a keyframe and ${\sim}14\%$ activate audio (Table~\ref{tab:perstep}), and
weighted by step count, ${\sim}61\%$ of steps are fully idle
(Figure~\ref{fig:obsactivity}). Audio-only families produce zero keyframes and
visual-only families produce zero audio activations, confirming clean channel separation.
Figure~\ref{fig:trajectories} traces this step-by-step on two representative tasks.

\begin{figure}[!h]
\centering
\begin{tcolorbox}[
  colback=white, colframe=black!50,
  title={\small\bfseries Meeting task --- ``What is the new launch date?''},
  fonttitle=\small, coltitle=white, colbacktitle=black!70,
  width=\textwidth, boxrule=0.5pt,
]
\small
\begin{minipage}[t]{0.47\textwidth}
\textbf{Standard (screenshot-only)} --- \textcolor{red}{\textbf{FAIL}}\\[2pt]
\textcolor{gray}{\scriptsize 15 steps, 51.2\,s}
\begin{enumerate}[leftmargin=*, label=\arabic*., itemsep=1pt]
  \item \texttt{wait()} \textcolor{gray}{\scriptsize --- sees static slide}
  \item \texttt{wait()} \textcolor{gray}{\scriptsize --- same slide}
  \item \texttt{wait()} \textcolor{gray}{\scriptsize --- same slide}
  \item \texttt{wait()} \textcolor{gray}{\scriptsize --- same slide}
  \item \texttt{wait()} \textcolor{gray}{\scriptsize --- same slide}
  \item[\vdots]
  \item[15.] \texttt{wait()} \textcolor{red}{\scriptsize \texttimes\ step budget exhausted}
\end{enumerate}
\vspace{4pt}
\textcolor{red}{\scriptsize Agent never hears the spoken date.}\\
\textcolor{red}{\scriptsize Waits all 15 steps without ever acting.}
\end{minipage}%
\hfill\vrule\hfill%
\begin{minipage}[t]{0.47\textwidth}
\textbf{AOI (full)} --- \textcolor{green!50!black}{\textbf{PASS}}\\[2pt]
\textcolor{gray}{\scriptsize 2 steps, 26.2\,s}
\begin{enumerate}[leftmargin=*, label=\arabic*., itemsep=1pt]
  \item \texttt{wait()}\\
    {\scriptsize \textbf{Audio:} ``Here's the team. As you can see we have strong cross-functional coverage.''}\\
    {\scriptsize \textbf{Narration:} ``Meeting slide shows Project Team with 6 members.''}
  \item \texttt{fill(``April 28th'')} \textcolor{green!50!black}{\scriptsize \checkmark}\\
    {\scriptsize \textbf{Audio:} ``We've moved the launch date to April 28th. Please update.''}\\
    {\scriptsize \textbf{Narration:} ``Audio announced the new launch date.''}
\end{enumerate}
\vspace{4pt}
\textcolor{green!50!black}{\scriptsize Agent hears ``April 28th'' and fills the form immediately.}
\end{minipage}
\end{tcolorbox}

\vspace{6pt}

\begin{tcolorbox}[
  colback=white, colframe=black!50,
  title={\small\bfseries Screencast task --- ``What was the first command in the terminal recording?''},
  fonttitle=\small, coltitle=white, colbacktitle=black!70,
  width=\textwidth, boxrule=0.5pt,
]
\small
\begin{minipage}[t]{0.47\textwidth}
\textbf{Standard} --- \textcolor{green!50!black}{\textbf{PASS}} (6 steps)\\[2pt]
\begin{enumerate}[leftmargin=*, label=\arabic*., itemsep=1pt]
  \item \texttt{wait()} \textcolor{gray}{\scriptsize --- recording playing}
  \item \texttt{wait()} \textcolor{gray}{\scriptsize --- missed first command}
  \item \texttt{wait()} \textcolor{gray}{\scriptsize --- scrolling through}
  \item \texttt{wait()}
  \item \texttt{wait()}
  \item \texttt{fill(\ldots)} \textcolor{green!50!black}{\scriptsize \checkmark\ lucky timing}
\end{enumerate}
\end{minipage}%
\hfill\vrule\hfill%
\begin{minipage}[t]{0.47\textwidth}
\textbf{AOI} --- \textcolor{green!50!black}{\textbf{PASS}} (1 step)\\[2pt]
\begin{enumerate}[leftmargin=*, label=\arabic*., itemsep=1pt]
  \item \texttt{fill(``git clone \ldots'')} \textcolor{green!50!black}{\scriptsize \checkmark}\\
    {\scriptsize \textbf{Narration:} ``The terminal recording shows the first command was git clone\ldots''}
\end{enumerate}
\vspace{4pt}
\textcolor{green!50!black}{\scriptsize Narration captures the command from the recording immediately.}
\end{minipage}
\end{tcolorbox}
\caption{Trajectory comparison on two representative tasks (\emph{illustrative, cherry-picked to expose the mechanism}).
\textbf{Top:} Meeting task where the launch date is spoken aloud.
The standard agent cannot hear it and exhausts its step budget on 15 \texttt{wait()} steps.
The AOI agent transcribes the audio, hears ``April 28th,'' and acts in 2 steps.
\textbf{Bottom:} Screencast task where the answer appears in a terminal recording.
The standard agent needs 6 steps with lucky timing. The AOI agent's narration captures the content in 1 step.}
\label{fig:trajectories}
\end{figure}

\begin{table}[!h]
\centering
\caption{Per-step observation statistics for Claude Sonnet~4.6 + AOI~full, by
family. \%\,KF = steps capturing $\geq1$ keyframe. KF/Step = mean keyframes/step.
\%\,Audio = steps with audio transcription.}
\label{tab:perstep}
\small
\begin{tabular}{@{}l rrrrr@{}}
\toprule
\textbf{Family} & \textbf{Steps} & \textbf{\%\,KF} & \textbf{Tot.\ KF} & \textbf{KF/Step} & \textbf{\%\,Audio} \\
\midrule
Podcast    & 35 & 8.6 & 3 & 0.09 & 42.9 \\
Meeting    & 34 & 50.0 & 18 & 0.53 & 73.5 \\
Screencast & 39 & 28.2 & 12 & 0.31 & 2.6 \\
Carousel   & 61 & 60.7 & 84 & 1.38 & 0.0 \\
Dashboard  & 52 & 28.8 & 18 & 0.35 & 0.0 \\
Transient  & 38 & 15.8 & 7 & 0.18 & 0.0 \\
Phone      & 38 & 0.0 & 0 & 0.00 & 26.3 \\
Interview  & 27 & 0.0 & 0 & 0.00 & 37.0 \\
Collab     & 39 & 10.3 & 4 & 0.10 & 17.9 \\
Games      & 118 & 36.4 & 65 & 0.55 & 0.0 \\
\midrule
\textbf{All} & \textbf{481} & \textbf{28.3} & \textbf{211} & \textbf{0.44} & \textbf{14.1} \\
\bottomrule
\end{tabular}
\end{table}

\section{Gemini-3 Flash Decomposition}
\label{app:gemini3}

This appendix provides the tables underlying the Gemini~3 case study (Figure~\ref{fig:gemini3}).
\begin{itemize}
  \item Table~\ref{tab:gemini3_audio_only} is the four-way decomposition behind the waterfall.
  \item Table~\ref{tab:gemini3_percat} is the per-family AOI~$\Delta$ (the shortfall vs.\ Gemini~2.5 sits entirely on Dashboard and Transient, where Gemini~3 alone regresses).
  \item Table~\ref{tab:keyframe_probe} is the causal probe that pins the keyframe regression to image-token dilution.
\end{itemize}

\begin{table}[!h]
\centering
\caption{Four-way decomposition of the Gemini~3 AOI residual. If the small net
were internal saturation, the audio-plus-standard-prompt row would not exceed the
full AOI. ``Aud4'' = the four audio families (Podcast, Meeting, Phone, Interview).
``D+T'' = Dashboard + Transient, where Gemini~3 regresses.}
\label{tab:gemini3_audio_only}
\small
\setlength{\tabcolsep}{4pt}
\begin{tabular}{@{}lcccccc@{}}
\toprule
\textbf{Mode} & \textbf{Total} & \textbf{Aud4} & \textbf{D+T} & \textbf{Scaffold?} & \textbf{Audio?} & \textbf{Keyframes?} \\
\midrule
Standard              & 36 & 10 & 11 & no  & no  & no  \\
Standard+scaffold     & 29 & 9  & 2  & yes & no  & no  \\
Standard+audio        & 48 & 20 & 7  & no  & yes & no  \\
\rowcolor{bestcolor}
Scaffold+audio (best) & \textbf{57} & 35 & 3  & yes & yes & no  \\
AOI full (default)    & 45 & 26 & 1  & yes & yes & yes \\
\bottomrule
\end{tabular}
\end{table}

\begin{table}[!h]
\centering
\caption{Gemini~3~Flash per-family success (pass/10), Standard vs.\ AOI~full. The
four audio families gain $+16$\,pp in aggregate. The shortfall vs.\ Gemini~2.5's
$+48$\,pp is concentrated in the Dashboard and Transient regressions.}
\label{tab:gemini3_percat}
\small
\setlength{\tabcolsep}{3pt}
\begin{tabular}{@{}l cccccccccc c@{}}
\toprule
\textbf{Mode} & Pod & Meet & Scr & Car & Dash & Tran & Phn & Int & Col & Gam & \textbf{Total} \\
\midrule
Standard & 0 & 2 & 4 & 4 & \textbf{8} & 3 & 1 & 7 & 5 & 2 & 36 \\
AOI full & \textbf{6} & \textbf{6} & \textbf{6} & \textbf{5} & 1 & 0 & \textbf{6} & \textbf{8} & 5 & 2 & 45 \\
\midrule
$\Delta$ & $+6$ & $+4$ & $+2$ & $+1$ & $-7$ & $-3$ & $+5$ & $+1$ & 0 & 0 & $+9$ \\
\bottomrule
\end{tabular}
\end{table}

\begin{table}[!h]
\centering
\caption{Keyframe causal probe on Gemini~3 (50 visual-active tasks), holding audio
and scaffold fixed and varying only the keyframe channel. Replacing keyframes with
same-size noise images reproduces the penalty, and the cost does not scale with
count or position, pointing to image-token dilution. Conditions are pairwise
indistinguishable ($p\geq0.375$ at $N{=}50$).}
\label{tab:keyframe_probe}
\small
\setlength{\tabcolsep}{4pt}
\begin{tabular}{@{}l cc l@{}}
\toprule
\textbf{Mode} & \textbf{Pass / 50} & \textbf{Rate} & \textbf{What it tests} \\
\midrule
No keyframes (anchor)        & \textbf{24} & 48\% & --- \\
Budget 1 keyframe            & 20 & 40\% & does count matter? \\
Budget 2 keyframes           & 18 & 36\% & \\
Budget 3 keyframes           & 20 & 40\% & \\
Default (budget 5)           & 18 & 36\% & \\
Noise images                 & 17 & 34\% & content vs.\ presence \\
Duplicate-screenshot images  & 20 & 40\% & content vs.\ presence \\
Keyframes after screenshot   & 22 & 44\% & position effect \\
\bottomrule
\end{tabular}
\end{table}

\section{DynaCU-Bench Task List}
\label{app:tasklist}

Each of the 100 tasks is identified by a category letter (A--J) and a difficulty-indexed ID: E1--E3 for easy tasks, M1--M4 for medium tasks, and H1--H3 for hard tasks.

Example tasks by category:
\begin{itemize}
  \item \textbf{A-E1} (Podcast, Easy): Listen to a podcast segment and identify the guest's name.
  \item \textbf{B-M2} (Meeting, Medium): During a meeting with slides, count the number of items in a presented chart.
  \item \textbf{C-H1} (Screencast, Hard): Watch a terminal recording and describe the full workflow demonstrated.
  \item \textbf{D-M4} (Carousel, Medium): Identify a specific product name from a rotating carousel.
  \item \textbf{E-E3} (Dashboard, Easy): Read the current warning level from a live-updating dashboard.
  \item \textbf{F-E2} (Transient UI, Easy): Click ``Accept'' on a cookie consent banner before it auto-dismisses.
  \item \textbf{G-M1} (Phone, Medium): Follow spoken instructions during a simulated phone call.
  \item \textbf{H-H2} (Interview, Hard): Complete a multi-question spoken interview with scoring.
  \item \textbf{I-M2} (Collab, Medium): Resolve a conflict in a collaborative document when a remote edit arrives.
  \item \textbf{J-E2} (Games, Easy): Complete 3 rounds of a simple reaction-time game.
\end{itemize}

All tasks are implemented as self-contained HTML files with embedded JavaScript for task logic, audio synthesis, and evaluation.
The complete benchmark (100 HTML files, evaluation scripts, and task definitions) is released publicly with the code.

\section{CLIP Threshold Calibration}
\label{app:theta}

The default threshold $\theta = 0.04$ was calibrated empirically.
Web UI events such as modal dialogs appearing produce cosine distances of ${\sim}0.08$. Video scene cuts produce distances of ${\sim}0.15$--$0.25$. Loading spinners and cursor blinks produce distances of ${\sim}0.005$--$0.02$.
Setting $\theta = 0.04$ captures all meaningful UI changes while filtering most periodic noise.

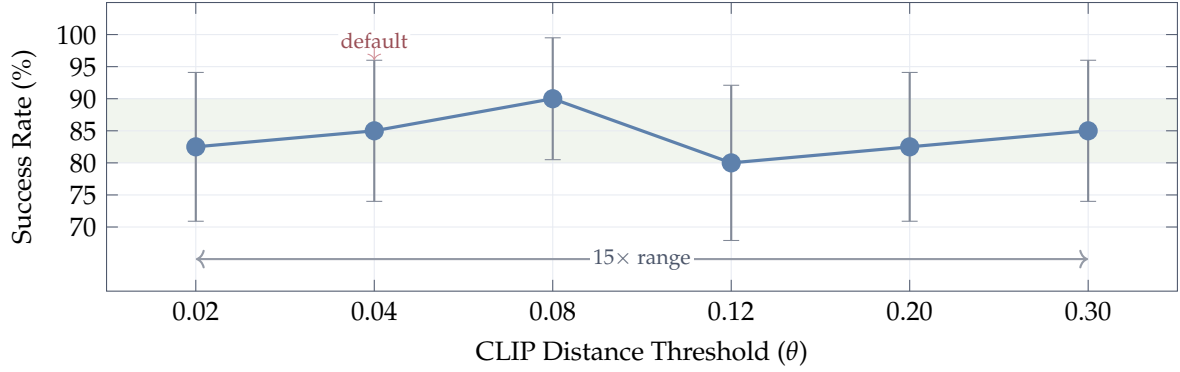
\begin{figure}[t]
\centering
\begin{tikzpicture}
\begin{axis}[
    width=\columnwidth,
    height=5.4cm,
    xlabel={CLIP Distance Threshold ($\theta$)},
    ylabel={Success Rate (\%)},
    xlabel style={font=\small},
    ylabel style={font=\small},
    xmin=0.5, xmax=6.5,
    ymin=60, ymax=105,
    xtick={1, 2, 3, 4, 5, 6},
    xticklabels={0.02, 0.04, 0.08, 0.12, 0.20, 0.30},
    xticklabel style={font=\small},
    ytick={70, 75, 80, 85, 90, 95, 100},
    yticklabel style={font=\small},
    grid=major,
    major grid style={cLightGray!60},
    axis line style={cGray},
    tick style={cGray},
    fill between/on layer={axis background},
]
\addplot[name path=upper, draw=none] coordinates {(0.4, 90) (6.6, 90)};
\addplot[name path=lower, draw=none] coordinates {(0.4, 80) (6.6, 80)};
\addplot[fill=cDelta!15] fill between[of=upper and lower];

\addplot[very thick, cAOI, mark=*, mark size=3pt, mark options={fill=cAOI},
         error bars/.cd, y dir=both, y explicit, error bar style={cGray!70, thick}]
    coordinates {
      (1, 82.5) +- (0, 11.6)
      (2, 85.0) +- (0, 11.0)
      (3, 90.0) +- (0, 9.5)
      (4, 80.0) +- (0, 12.1)
      (5, 82.5) +- (0, 11.6)
      (6, 85.0) +- (0, 11.0)
    };

\node[font=\scriptsize, text=cStandard!80!black, fill=white, inner sep=1pt, rounded corners=1pt]
    at (axis cs:2, 99) {default};
\draw[->, thin, cStandard!70] (axis cs:2, 98) -- (axis cs:2, 96);

\draw[<->, thick, cGray!60] (axis cs:1, 65) -- (axis cs:6, 65);
\node[font=\scriptsize, text=cGray, fill=white, inner sep=1pt] at (axis cs:3.5, 65) {$15\times$ range};

\end{axis}
\end{tikzpicture}
\caption{\textbf{CLIP threshold ($\theta$) sensitivity} on 40 visual tasks (categories C--F) with Claude Sonnet~4.6.
Error bars are 95\% Wilson confidence intervals at $N{=}40$ ($\pm$9.5--12.1\,pp).
Point estimates remain within an 80--90\% band (shaded) across a $15\times$ range of $\theta$, but the CIs overlap heavily, so we cannot distinguish individual thresholds at this sample size.
$\theta$ values are placed at uniform spacing on the axis (not to scale).
The bimodal distribution of CLIP distances (near-zero for unchanged content, large for meaningful changes) makes the method naturally insensitive to $\theta$.}
\label{fig:theta}
\end{figure}

Figure~\ref{fig:theta} shows the CLIP threshold $\theta$ sweeps across a 15$\times$ range from 0.02 to 0.30, evaluated on the 40 visual tasks (categories C--F) where keyframe selection is relevant.
Point estimates remain within an 80--90\% band throughout, demonstrating that for practitioners who choose the CLIP-based variant, threshold selection is not critical.
There is no cliff: even at $\theta = 0.30$ (very aggressive filtering that captures only major visual changes), the point estimate remains at 85\%.
At $\theta = 0.02$ (very sensitive, capturing small changes), it is 82.5\%.
The point estimate peaks at $\theta = 0.08$ (90\%), but with $N{=}40$ the 95\% Wilson confidence intervals on every point span $\sim$10--12\,pp and overlap heavily. We therefore do not claim a true sweet spot, only that the method is robust across a $15\times$ range of $\theta$ and that any reasonable choice will work.
This insensitivity arises because CLIP embeddings produce a bimodal distribution of distances: semantically unchanged frames cluster near zero, while meaningful changes produce large distances, with relatively few samples in between.

\section{Glossary of Observation Modes}
\label{app:modes}

\begin{table}[!h]
\centering
\caption{Every observation mode used in the paper.
KF = inter-step keyframes (with selection strategy).
Audio = transcripts of system audio.
Narr.\ = CU-model visual narration persisted as text.
Elem.\ = DOM element list in the prompt.
Traj.\ = structured prior-step trajectory wrapping.
\textbf{Standard} and \textbf{AOI full} run on all models. The rest are
Claude Sonnet~4.6 ablations except where a section indicates a Gemini~3 or
open-source diagnostic.}
\label{tab:modes}
\footnotesize
\setlength{\tabcolsep}{3pt}
\begin{tabular}{@{}l ccccc l@{}}
\toprule
\textbf{Mode} & \textbf{KF} & \textbf{Audio} & \textbf{Narr.} & \textbf{Elem.} & \textbf{Traj.} & \textbf{Where} \\
\midrule
Standard & --- & --- & --- & --- & \checkmark & Table~\ref{tab:main} \\
Minimal prompt & --- & --- & --- & --- & --- & App.~\ref{app:promptdecomp} \\
$+$element list & --- & --- & --- & \checkmark & --- & App.~\ref{app:promptdecomp} \\
$+$format (scaffold) & --- & --- & --- & \checkmark & \checkmark & \S\ref{sec:promptperc} \\
Standard$+$audio & --- & \checkmark & --- & --- & \checkmark & App.~\ref{app:gemini3} \\
\midrule
Selection variants & variant & --- & --- & \checkmark & \checkmark & \S\ref{sec:ablation} \\
$+$keyframes (visual) & CLIP & --- & --- & \checkmark & \checkmark & \S\ref{sec:ablation} \\
$+$ASR & CLIP & \checkmark & --- & \checkmark & \checkmark & \S\ref{sec:ablation} \\
Scaffold$+$audio (no KF) & --- & \checkmark & \checkmark & \checkmark & \checkmark & \S\ref{sec:keyframe_context} \\
\textbf{AOI full} & pixel gate & \checkmark & \checkmark & \checkmark & \checkmark & Table~\ref{tab:main} \\
Narration discarded & pixel gate & \checkmark & not kept & \checkmark & \checkmark & \S\ref{sec:narration_discard} \\
Keyframe probes & modified & \checkmark & \checkmark & \checkmark & \checkmark & App.~\ref{app:gemini3} \\
\bottomrule
\end{tabular}
\end{table}

\section{Implementation Details}
\label{app:implementation}

\textbf{Software.}
Python 3.11, Playwright 1.49 for browser automation, PulseAudio 16.1 for audio I/O, edge-TTS for high-quality speech synthesis, OpenAI CLIP ViT-B/16 for keyframe extraction, Whisper large-v3 for speech transcription, vLLM 0.19.0 for local model serving.

\textbf{Hardware.}
All experiments run on a single machine with an NVIDIA RTX PRO 6000 Blackwell GPU (96\,GB VRAM), 192\,GB RAM.
Cloud models (Claude, GPT-5.4, Gemini~2.5 Flash) are accessed via HTTP APIs.
Local models (EvoCUA-32B, Fara-7B) are served via vLLM with 4-bit quantization (\texttt{bitsandbytes}).
Whisper runs on CPU to avoid GPU contention.
Running the two local models in the \texttt{standard\_structured} mode
(Section~\ref{sec:promptperc}) required patching vLLM~0.19.0 around an
NVML/userspace driver mismatch on our host: the CUDA platform plugin
trusts \texttt{pynvml.nvmlInit()} alone, and an overly strict
free-memory assertion fired when the Whisper service released GPU
memory during startup profiling.  Both patches are released with the
code.

\textbf{Hyperparameters.}
Screen capture rate: ${\sim}3$\,Hz.
Pixel change threshold ($\alpha$): 1\%.
CLIP distance threshold ($\theta$): 0.04.
Maximum keyframes per step: 5.
Audio capture: a 70\,s ring buffer at 16\,kHz mono holds rolling history.
When the volume gate fires, Whisper is invoked once per step on the new
inter-step audio (typically ${\sim}3.5$\,s) extended by a ${\sim}3.5$\,s
overlap into the previous interval (${\sim}7$\,s total) for boundary
continuity.
Post-action wait: 2.0\,s.
Maximum steps per task: 15.

\section{DynaCU-Real-Local Details}
\label{app:realcontent}

The 12 tasks span four sub-domains: 3~podcast (Aesop animal-identification),
3~meeting (Python / Postgres / Rust technical detail), 3~screencast (\texttt{git
clone} / \texttt{pip install} / \texttt{npm test} outcome), and 3~voice
(yes/no / directions / appointment day).
Audio for podcast/meeting/voice tasks is rendered with \texttt{espeak}
(a formant synthesizer with a deliberately non-neural acoustic profile).
Screencast tasks use real \texttt{asciinema} v2 cast files generated locally.
Source texts are public-domain materials: Aesop's Fables for podcasts,
Wikipedia for Python language facts, project documentation for
PostgreSQL/MVCC and Rust/tokio.
All HTML harnesses, asset-build scripts, and success checks are released
with the benchmark.

\begin{table}[t]
\centering
\caption{DynaCU-Real-Local: Claude~Sonnet~4.6 Standard vs.\ AOI~full on 12
real-recording tasks.  Both modes tie at $11/12$ on this set. The failure
is the same task (\texttt{R\_cast2\_pip}) for both.}
\label{tab:realcontent}
\small
\begin{tabular}{@{}l cccc c@{}}
\toprule
\textbf{Mode} & \textbf{Podcast} & \textbf{Meeting} & \textbf{Screencast} & \textbf{Voice} & \textbf{Total / 12} \\
\midrule
Standard      & 3/3 & 3/3 & 2/3 & 3/3 & \textbf{11} \scriptsize(92\%) \\
AOI full      & 3/3 & 3/3 & 2/3 & 3/3 & \textbf{11} \scriptsize(92\%) \\
\bottomrule
\end{tabular}
\end{table}

\section{Streaming-Baseline Adapter Sanity Check Details}
\label{app:streaming_sanity}

To rule out the explanation that the streaming baselines' weak
audio-subset numbers (Section~\ref{sec:streaming}) are an artifact of our
adapter rather than a model limitation, we evaluate both adapters on a
small purely-visual sanity-check set of five tasks across four categories: C
(video / static recording), E (live dashboard), F (transient UI, with two
tasks: one banner-accept and one cookie-accept subtype), and I
(collaborative document).
All categories are chosen such that the answer can be reached without any audio comprehension.
The exact set is C-E1, E-E1, F-E1, F-E2, and I-E1.
For each baseline, we open one streaming session per task, send the
post-action screenshot as a multimodal turn, and translate any returned
tool calls into browser actions through the same path used in the main
audio-subset evaluation.
The success criterion is the same DOM check as in the main benchmark.
Table~\ref{tab:streaming_sanity_detail} reports the per-task outcome and
the modal failure mode where applicable.

\begin{table}[t]
\centering
\caption{Per-task adapter sanity-check outcomes for each streaming
baseline.  Every adapter drives the browser and emits valid tool calls
(\texttt{click}, \texttt{fill}, \texttt{type\_text}) on every task. Every
attempted action lands on the DOM, so the audio-subset failures are
content-driven rather than infrastructural.  The older adapters score 0/5
task-correct, failing by wrong target (e.g.\ F-E2 cookies dismissed instead of accepted),
hallucinated content (e.g.\ C-E1), or premature commit (e.g.\ I-E1), whereas the
current GA \texttt{gpt-realtime-2} reaches 3/5, closer to but still below the
AOI reference, consistent with the audio subset (Table~\ref{tab:streaming}).}
\label{tab:streaming_sanity_detail}
\small
\setlength{\tabcolsep}{3.5pt}
\begin{tabular}{@{}l ccccc c@{}}
\toprule
\textbf{Baseline} & \textbf{C-E1} & \textbf{E-E1} & \textbf{F-E1} & \textbf{F-E2} & \textbf{I-E1} & \textbf{Total} \\
\midrule
OpenAI Realtime adapter (gpt-4o backbone) & \ding{55} & \ding{55} & \ding{55} & \ding{55} & \ding{55} & 0 \\
Gemini Live (2.5 flash native-audio)    & \ding{55} & \ding{55} & \ding{55} & \ding{55} & \ding{55} & 0 \\
OpenAI \texttt{gpt-realtime-2} (GA, native audio) & \checkmark & \checkmark & \checkmark & \ding{55} & \ding{55} & 3 \\
\rowcolor{bestcolor}
AOI~full (Claude Sonnet 4.6) reference   & \checkmark & \ding{55} & \checkmark & \checkmark & \checkmark & 4 \\
\bottomrule
\end{tabular}
\end{table}

\end{document}